\pdfoutput=1

\documentclass[11pt]{article}

\usepackage{bm}
\usepackage[preprint]{acl}
\usepackage{colortbl}
\usepackage{multirow}
\usepackage{times}
\usepackage{rotating}
\usepackage[normalem]{ulem}
\usepackage{latexsym}
\usepackage{booktabs} 
\newcommand{\hzz}[1]{{\color{red}{#1}}}
\usepackage[T1]{fontenc}
\usepackage{amsmath}
\usepackage{amssymb}
\usepackage[utf8]{inputenc}
\usepackage{float}
\usepackage{microtype}
\usepackage{enumitem}
\usepackage{inconsolata}
\usepackage{mdframed}
\usepackage{graphicx}

\title{\raisebox{-1.5ex}{\includegraphics[width=0.09\textwidth]{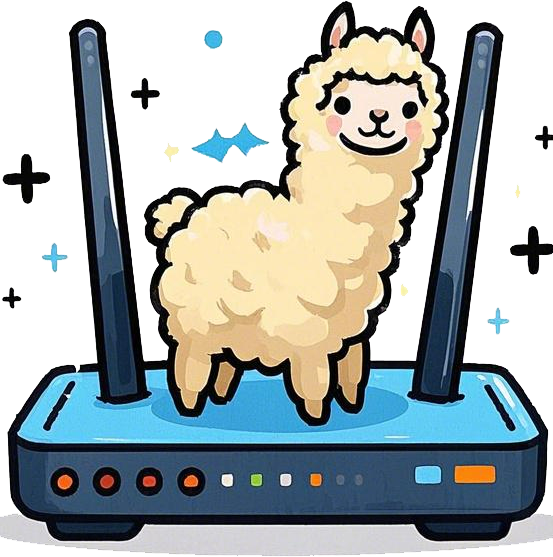}} RouterEval: A Comprehensive Benchmark for Routing LLMs to Explore Model-level Scaling Up in LLMs}

\author{
 \textbf{Zhongzhan Huang\textsuperscript{1}},
 \textbf{Guoming Ling\textsuperscript{1}},
 \textbf{Yupei Lin\textsuperscript{1}},
 \textbf{Yandong Chen\textsuperscript{1}},
 \\
  \textbf{Shanshan Zhong\textsuperscript{1}},
  \textbf{Hefeng Wu\textsuperscript{1}},
   \textbf{Liang Lin\textsuperscript{1}}
\\
 \textsuperscript{1}Sun Yat-sen University
\\
}

\begin{document}
\maketitle
\begin{abstract}
Routing large language models (LLMs) is a new paradigm that uses a router to recommend the best LLM from a pool of candidates for a given input.
In this paper,
our comprehensive analysis with more than 8,500 LLMs reveals a novel model-level scaling up phenomenon in Routing LLMs, i.e., a capable router can significantly enhance the performance of this paradigm as the number of candidates increases.
This improvement can even surpass the performance of the best single model in the pool and many existing strong LLMs, confirming it a highly promising paradigm. 
However, the lack of comprehensive and open-source benchmarks for Routing LLMs has hindered the development of routers. In this paper, we introduce RouterEval 
\footnote{See \href{https://github.com/MilkThink-Lab/RouterEval}{project page} 
for all \hzz{\textbf{data, code and tutorial}}.}
, a benchmark tailored for router research, which includes over 200,000,000 performance records for 12 popular LLM evaluations across various areas such as commonsense reasoning, semantic understanding, etc., based on over 8,500 various LLMs. 
Using RouterEval, extensive evaluations of existing Routing LLM methods reveal that most still have significant room for improvement. 
\end{abstract}

\section{Introduction}
\begin{figure}[t]
  \includegraphics[width=\columnwidth]{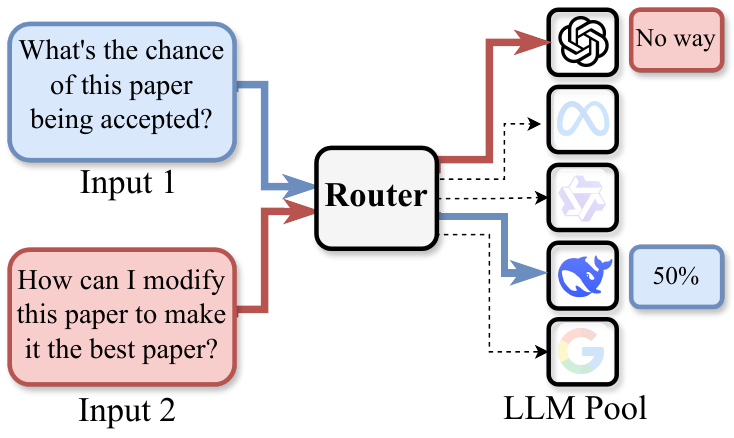}
  \caption{\textbf{The Overview of Routing LLMs}. For each given input, the router distributes it to the appropriate LLM  to achieve specific objectives, such as high accuracy, low computational cost, reduced hallucinations, etc. We find that Routing LLMs is a promising paradigm for LLMs to achieve model-level scaling up.
}
  \label{fig:router}
  \vspace{-5pt}
\end{figure}

Routing LLMs~\cite{jitkrittum2025universal,lu2023routing,zhao2024model,shnitzer2023large,chen2024routerdc} aims to establish an efficient router capable of selecting an LLM from a pool of candidates with varying abilities to effectively handle a given input, in order to achieve specific goals such as high overall accuracy, low computational cost, or minimal hallucination, as shown in Fig.~\ref{fig:router}. Since this paradigm only involves input assignment, it is naturally compatible with heterogeneous LLMs of different structures, as well as most model enhancement methods, including fine-tuning~\cite{hu2022lora,zhang2023lora}, Mixture-of-Experts~\cite{jiang2024mixtral,zhong2024moextend}, model merging~\cite{goddard2024arcee,yang2024model}, etc. This allows for the creation of a sufficiently large LLM candidate pool. In this paper, through extensive experiments, we find a model-level scaling up phenomenon in LLMs, i.e., using some capable router, the rapid performance improvement as the number of candidates in the LLM pool increases, can even easily surpass the performance of the best single model in the pool and most existing strong LLMs. Moreover, the majority of LLMs in the candidate pool are relatively small and open-source models whose individual capabilities are far below those of commercialized LLMs, 
like GPT-4~\cite{achiam2023gpt}, which demonstrates Routing LLM is a highly promising paradigm (See Section \ref{sec:poten} for details).

\begin{figure*}[t]
 \centering
  \includegraphics[width=0.9\linewidth]{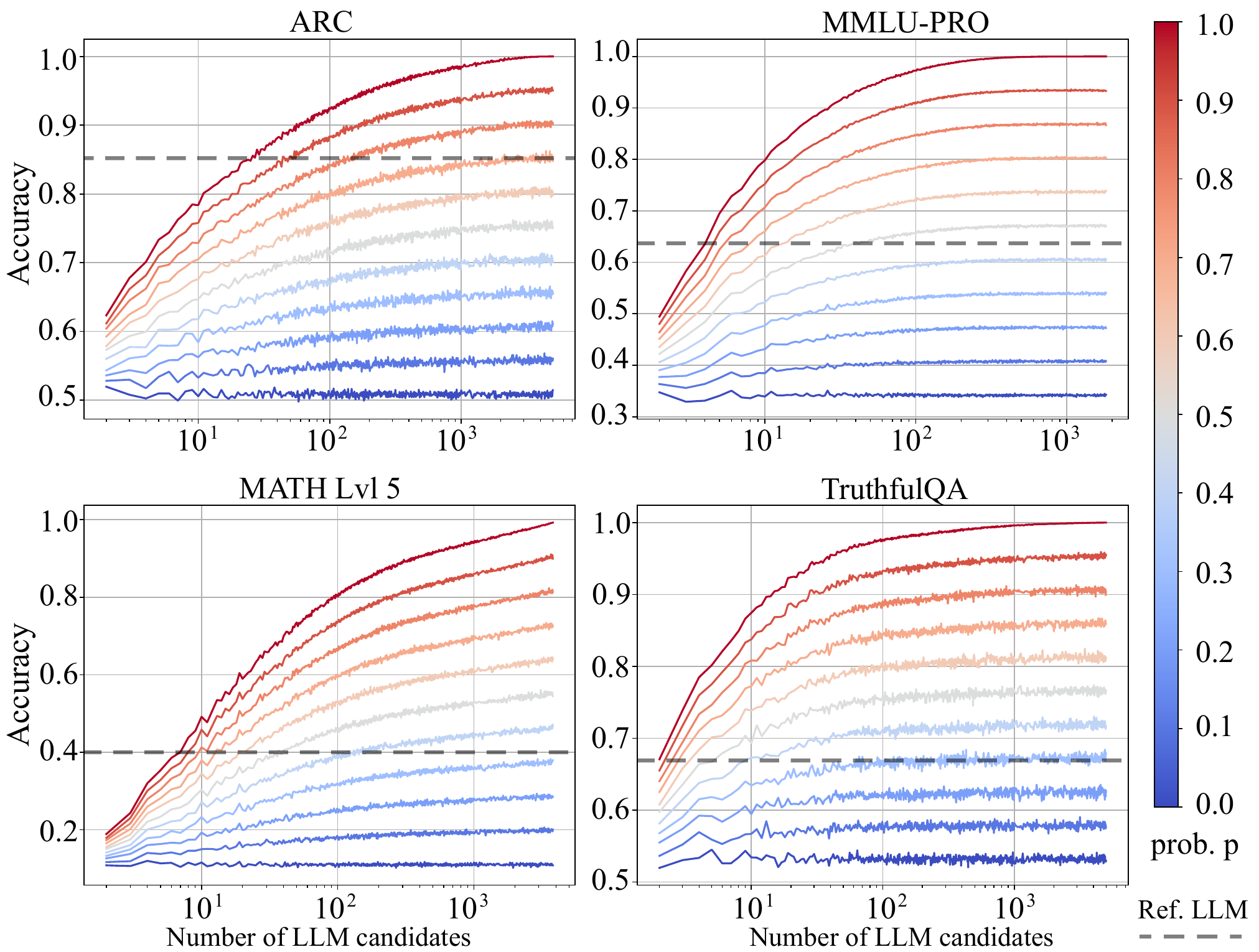}
  \caption{\textbf{The Model-level Scaling Up Phenomenon in Routing LLMs}.   As shown in Section \ref{sec:poten}, the Prob. $p$ indicates the performance of the router, with values closer to 1 representing greater similarity to the oracle router's capability. If $p \to 0$, then $r_o(p)$ degenerates into a random sampler. When the router $r_o(p)$ reaches a certain level of capability, it induces a scaling up phenomenon in the Routing LLMs paradigm. Specifically, as the number of LLM candidates increases, performance rapidly improves. "Ref. LLM" denotes a representative LLM with strong performance on given benchmark, such as GPT-4. Further details are provided in Section \ref{sec:metrics}. For more examples of model-level scaling up phenomenon, please refer to the Appendix \ref{sec:more}.}
  \label{fig:scaling}
\end{figure*}
However, current Routing LLMs methods are still in the early stages of rapid development and lack comprehensive and open-source benchmarks specifically designed for routers, which hinders their progress. For example, some existing benchmarks~\cite{lu2024merge,hu2024routerbench} still have the issues of  insufficient LLM candidates, limited evaluation diversity, closed-source or inadequate performance records, etc. To address these challenges, in this paper, we collect and clean 12 popular LLM evaluations covering fields such as knowledge-based Q\&A, commonsense inference, semantic understanding, etc. From these evaluations, we use performance records from over 8,500 LLMs, amounting to more than 200,000,000 entries, to construct a comprehensive Router benchmark named RouterEval in Section \ref{sec:benchmark}.
Utilizing RouterEval, in Section \ref{sec:exp}, we conduct a comprehensive exploration of several existing router construction methods across various settings and find that these methods still have considerable room for improvement.
We show the related works in Appendix \ref{sec:related}, and summarize the contributions of this paper as follows:
\begin{itemize}[leftmargin=15pt]
     \item We find the model-level scaling up phenomenon in LLMs, i.e., capable router in Routing LLMs can significantly enhance the performance  as the number of candidates increases. 
     \vspace{-5pt}
     \item Based on over 8,500 LLMs and their performance records exceeding 200 million entries across various LLM evaluations, we construct a comprehensive benchmark tailored for router design, named RouterEval, to re-examined a wide range of existing router methods.
\end{itemize}

\section{Preliminary}
\label{sec:pre}
\noindent\underline{\textbf{Notation.}} Assume there is a LLM pool containing \( m \) LLMs \(\{\ell_i\}_{i=1}^m\) and input set $\{s_j\}_{j=1}^n$. For a given input \( s_j  \), its representation \(\kappa(s_j)\) can be obtained using an encoder \(\kappa\). Based on the performance records of these LLMs on \( s_j \), we can obtain a \( m \)-dimensional one-hot selection vector \( v_j \), where the dimension marked as 1 indicates that the corresponding LLM indexed by that dimension achieves the dominant performance among \(\{\ell_i\}_{i=1}^m\) on the input \( s \), otherwise it is marked as 0. The vector \( v_j \) indicates the optimal choice for the router on the input \( s_j \). 
Specifically, taking $m=3$ as example, if the performance metric is correctness or incorrectness,  the dimensions of \( v_0 \) corresponding to the LLMs that provide correct answers can be simultaneously marked as 1, such as \( v_0=[1,1,0] \) or even \( v_0=[1,1,1] \), since these LLMs can all be considered as the optimal choices for the router. Moreover, If the metric is continuous, the index of the LLM whose performance is within 95\% of the optimal performance can be marked as 1, otherwise, it is marked as 0.

\noindent\underline{\textbf{Router.}} 
In practice, the task of the router $r_\theta$ with learnable parameters $\theta$ is to establish the following mapping using the given input set \(\{s_j\}_{j=1}^n\) and the corresponding selection vectors \(\{v_j\}_{j=1}^n\), and possible external data $\mathcal{D}$ as the training set:
\begin{equation}
    r_\theta[\kappa(s_j)|\mathcal{D}] \to v_j.
    \label{eq:obj}
\end{equation}
If there is no external data, then $\mathcal{D} = \phi$. After training, for an unseen input \( s' \), the trained router is required to generate a selection vector \( v' \), and based on this vector, determine the optimal LLM from the LLM pool for the current input. Thus, the acquisition of the router can be regarded as a classic classification problem~\cite{deng2009imagenet,hu2018squeeze,zhong2023lsas,zhong2023esa,liang2020instance,huang2020dianet}.

Given that the development of routers is still in its nascent stage, in this paper, we prioritize the performance corresponding to each benchmark when selecting LLMs, without considering factors such as computational cost, hallucination rate, etc. See Section \ref{sec:ana} for more analysis.

\section{Model-level Scaling Up Phenomenon}
\label{sec:poten}

In this section, we comprehensively explore the relationship between the number of LLM candidates and the final performance under different router capabilities, to demonstrate the immense potential of Routing LLMs paradigm. 
Specifically, we consider the four well-known LLM benchmarks ARC~\cite{clark2018think}, MMMU-PRO~\cite{wang2024mmlu}, MATH Lvl 5~\cite{hendrycks2021measuring}, TruthfulQA~\cite{lin2021truthfulqa}, and collect performance records of thousands of LLMs on these benchmarks. 
Using collected performance records, we can construct an oracle router \( r_o \), which can select the optimal LLM to process a given input \( s \) from any \( m \) LLM candidates. Furthermore, we can construct other routers with different capabilities by defining \( r_o(p) \) as
\begin{equation}
    r_o(p) = \begin{cases}
r_o, & \text{with probability } p, \\
\omega_m, & \text{with probability } 1-p,
\end{cases}
\end{equation}

where \( \omega_m \) is a router that samples uniformly from the \( m \) candidate LLMs with probability \( 1/m \). As \( p \to 1 \), \( r_o(p) \) approaches the oracle router \( r_o \), resulting in strongest classification performance on $m$ LLM candidates. Conversely, as \( p \to 0 \), \( r_o(p) \) degenerates into a random sampler.

Subsequently, for a given \( m \), we repeatedly sample \( m \) LLMs from the  LLM pool with equal probability, 100 times, forming 100 candidate subsets each containing \( m \) LLM candidates. We then calculate the average performance of different routers \( r_o(p) \) on these subsets. The statistical results are shown in Fig.~\ref{fig:scaling}, and we have following finding:

\begin{mdframed}[backgroundcolor=gray!8]
\begin{minipage}{\linewidth}
\vspace{2pt}
(1) Model-level Scaling Up for LLMs.
\vspace{2pt}
\end{minipage}
\end{mdframed}
 As the number of candidates increases, most routers $r_o(p)$ can rapidly enhance performance, especially when $p \geq 0.5$. This implies that, given sufficiently well-constructed routers, Routing LLM is an effective paradigm for scaling up LLMs at the model-level. In fact, this phenomenon aligns with the classical neural scaling law~\cite{kaplan2020scaling}, as the increase in the number of candidates can be regarded as an implicit increase in the number of parameters, which is manifested in a sparse manner under the Routing LLM paradigm.

\begin{mdframed}[backgroundcolor=gray!8]
\begin{minipage}{\linewidth}
\vspace{2pt}
(2) Weak Candidates Can Also be Promising.
\vspace{2pt}
\end{minipage}
\end{mdframed}
In fact, the vast majority of LLMs discussed in this paper are open-source and can be deployed locally (see Appendix \ref{sec:details} for details). The individual performance of these models is not particularly remarkable and generally falls significantly short of mainstream LLMs. However, as shown in Fig.~\ref{fig:scaling}, even relatively weak candidates can achieve complementary performance among multiple heterogeneous models under the Routing LLMs paradigm. They can obtain fine-grained division of labor for each input and outperform mainstream LLMs. This shows that relatively weak candidates can also be promising for obtaining high performance within this paradigm. See more details in Section \ref{sec:constr}.

\begin{mdframed}[backgroundcolor=gray!8]
\begin{minipage}{\linewidth}
\vspace{2pt}
(3) Small Number of Candidates is Enough.
\vspace{2pt}
\end{minipage}
\end{mdframed}
In Fig.~\ref{fig:scaling}, we not only observe that a sufficiently large number of candidates can achieve excellent performance, but also note that even with only 3 $\sim$ 10 candidates, the performance can obtain strong performance, and even surpass that of the strong reference model. Therefore, when only moderately good performance is required, a small number of candidates is sufficient, which is highly advantageous for users with limited resources to adopt the Routing LLMs paradigm. Coupled with the aforementioned observation that weak candidates also hold great potential, the Routing LLMs paradigm demonstrates significant utility and applicability.

\section{The Construction of RouterEval}
\label{sec:benchmark}

In this section, we elaborate on the construction of our comprehensive benchmark for Routing LLMs, named \textbf{RouterEval}. We collected over 200,000,000 performance records for 12 popular LLM evaluations spanning areas such as knowledge-based Q\&A, commonsense reasoning, semantic understanding, and instruction following, based on more than 8,500 LLMs. The benchmarks~\cite{zellers2019hellaswag,clark2018think,wang2024mmlu,hendrycks2020measuring,lin2021truthfulqa,sakaguchi2021winogrande,cobbe2021training,zhou2023instruction,suzgun2022challenging,rein2024gpqa,sprague2023musr} involved include ARC, HellaSwag, MMLU, TruthfulQA, Winogrande, GSM8k, IFEval, BBH, GPQA, MUSR, MATH Lvl 5, and MMLU-PRO. See Appendix \ref{sec:details}, \ref{sec:detai_routereval} and \ref{sec:details_group} for more details.

\subsection{Data format}
\label{sec:format}
As shown in Eq.~(\ref{eq:obj}), our objective is to train a router \( r_\theta \) with learnable parameters \( \theta \), which is essentially an \( m \)-way classifier. Given the representation \( \kappa(s_j) \) of an input \( s \), the router fits a selection vector \( v_j \in \{0,1\}^m \) and identifies the optimal LLM for the input \( s \), corresponding to the position of 1 in \( v_j \). Accordingly, the input $\mathcal{X}$ and label $\mathcal{Y}$ format of RouterEval is 
\begin{equation}
  (\mathcal{X},\mathcal{Y}) = \{\kappa(s_j), v_j\}_{j=1}^n,
  \label{eq:format}
\end{equation}
 where \( s_j \) encompasses all test samples from the 12 LLM benchmarks considered in this study. We also provide versions using four different pre-trained models~\cite{beltagy2020longformer,reimers2019sentence,liu2019roberta} as encoders $\kappa$: Sentence BERT, RoBERTa, the last layer of RoBERTa, and Longformer. The training, validation, and test sets are split in a ratio of 8:1:1. In all experiments, Roberta is used as an example. Researchers can select any embedding in our code or design their own embedding based on input.

\subsection{The Construction of LLM Candidates}
\label{sec:constr}
The selection vector \( v_j \in \{0,1\}^m \) in Eq.~(\ref{eq:format}) depends on the choice of \( m \) LLM candidates. In this paper, we set two difficulty levels: an easy level with \( m \in \{3,5\} \) and a hard level with \( m \in \{10, 100,1000\} \). We focus primarily on the easy level, as the analysis in Fig.~\ref{fig:scaling} and Section \ref{sec:poten} shows that performance grows most rapidly when \( 2 \leq m \leq 10 \). Moreover, the deployment cost of Routing LLMs is low in this range, making these values of \( m \) the most cost-effective and worthy of attention. The hard level of \( m \) is mainly used to explore the limits of router design, in preparation for the strong performance demonstrated by the scaling up phenomenon in Section \ref{sec:poten}.

For each given benchmark and \( m \), we construct three types of LLMs candidate $G$, and the final model performance is the average of the results from these three candidates. Specifically, we sort all \( N \) LLMs with performance between 0.1 and 0.9 based on their individual performance on the given benchmark, obtaining \( \{\ell_i^\prime\}_{i=1}^N \). We then consider the following optimization problem regarding the performance of \( G \) and its corresponding oracle \( r_o \), as shown in Eq.~(\ref{eq:opti}),
\begin{equation}
 \hat{G} = \max\nolimits_G \text{Perf.}(r_o,G).
    \label{eq:opti}
\end{equation}
When the \( m \) LLMs in \( G \) are all selected from \( \{\ell_i^\prime\}_{i=1}^{\lfloor0.2N\rfloor} \) and \( \{\ell_i^\prime\}_{i=\lfloor0.8N\rfloor}^{N} \) respectively, \( \hat{G} \) forms the "all-strong" group and the "all-weak" group. And \( \hat{G} \) forms the "strong-to-weak" group, when the \( j \)-th LLM in \( G \) is selected from \( \{\ell_i^\prime\}_{i=(j-1)m}^{\min(jm,N)} \).
Through this process, we can explore the router from multiple perspectives and investigate the potential of Routing LLMs. For instance, in MMLU and $m=10$, while the individual performance of each LLM in the "all-weak" group does not exceed 0.3, its oracle performance can reach 0.95. See Appendix \ref{sec:details_group} for more details. Additionally, through validation, we ensure that in all candidate groups, at least 
$10^{30}$
 routers can achieve oracle performance, showing the learnability of the routers.

\subsection{Extra Training Data}
\label{sec:extra}
Note that, the direct training set in RouterEval typically ranges from several hundred to tens of thousands. This poses a significant challenge for training a reliable router, especially for the hard level \( m \) discussed in Section \ref{sec:constr}. Therefore, similar to what is shown in Eq.~(\ref{eq:obj}), we provide an extra dataset \( \mathcal{D} \) to aid in the training of the router. Specifically, we open-source over 200,000,000 performance records constructed from various LLMs and benchmarks involved in this study. Utilizing these data, researchers can explore various data augmentation techniques~\cite{shorten2019survey,qin2024introspective}, few-shot learning~\cite{vinyals2016,huang2024attns}, regularization method~\cite{srivastava2014,huang2021altersgd,huang2023scalelong}, pre-training approaches~\cite{zoph2020rethinking,zhong2024let}, and  recommendation systems~\cite{zhao2025dvib,zhong2024mirror}, etc.  Additionally, to further enhance the capabilities of the router, our proposed RouterEval encourages the use of any external data and pre-trained models.

\begin{table*}[t]
  \centering

  \resizebox*{0.99\linewidth}{!}{
\begin{tabular}{|c|l|rrrr|rrrr|rrrr|rrrr|}
    \toprule
    \multicolumn{1}{|r}{} &       & \multicolumn{4}{c|}{ARC}      & \multicolumn{4}{c|}{HellaSwag} & \multicolumn{4}{c|}{MMLU}     & \multicolumn{4}{c|}{TruthfulQA} \\
\cmidrule{3-18}    \multicolumn{1}{|r}{} & Router & \multicolumn{1}{l}{$\mu_o$↑} & \multicolumn{1}{l}{$V_R$↑} & \multicolumn{1}{l}{$V_B$↑} & \multicolumn{1}{l|}{$E_p$} & \multicolumn{1}{l}{$\mu_o$↑} & \multicolumn{1}{l}{$V_R$↑} & \multicolumn{1}{l}{$V_B$↑} & \multicolumn{1}{l|}{$E_p$} & \multicolumn{1}{l}{$\mu_o$↑} & \multicolumn{1}{l}{$V_R$↑} & \multicolumn{1}{l}{$V_B$↑} & \multicolumn{1}{l|}{$E_p$} & \multicolumn{1}{l}{$\mu_o$↑} & \multicolumn{1}{l}{$V_R$↑} & \multicolumn{1}{l}{$V_B$↑} & \multicolumn{1}{l|}{$E_p$} \\
   \midrule
    \multirow{8}[1]{*}{\begin{sideways}$m=3$\end{sideways}} & \cellcolor[rgb]{ .996,  .89,  .871}Oracle $r_o$ & \cellcolor[rgb]{ .996,  .89,  .871}0.80  & \cellcolor[rgb]{ .996,  .89,  .871}0.94  & \cellcolor[rgb]{ .996,  .89,  .871}1.34  & \cellcolor[rgb]{ .996,  .89,  .871}1.02  & \cellcolor[rgb]{ .996,  .89,  .871}0.80  & \cellcolor[rgb]{ .996,  .89,  .871}0.84  & \cellcolor[rgb]{ .996,  .89,  .871}1.08  & \cellcolor[rgb]{ .996,  .89,  .871}1.32  & \cellcolor[rgb]{ .996,  .89,  .871}0.89  & \cellcolor[rgb]{ .996,  .89,  .871}1.03  & \cellcolor[rgb]{ .996,  .89,  .871}1.35  & \cellcolor[rgb]{ .996,  .89,  .871}1.00  & \cellcolor[rgb]{ .996,  .89,  .871}0.85  & \cellcolor[rgb]{ .996,  .89,  .871}1.27  & \cellcolor[rgb]{ .996,  .89,  .871}1.21  & \cellcolor[rgb]{ .996,  .89,  .871}1.05  \\
\end{tabular}%
    }
      \caption{\textbf{The Results on RouterEval (part1)}.See Section \ref{sec:metrics} for detials of various metrics.  Red area and blue area highlights indicate the "Strong router" and "Existing router" mentioned in Section \ref{sec:baseline}, respectively. The best results in existing router methods are highlighted with underlines and on bold. The values in the table are rounded to two decimal places. The results on hard level settings are shown in Appendix \ref{sec:res_hard}.}
  \label{tab:res1}%
\end{table*}%

\section{Experiments}
\label{sec:exp}
In this section, we present the performance of various existing routers on our proposed RouterEval. Specifically, in these experiments, we do not consider the extra data mentioned in Section \ref{sec:extra}, since the utilization of such data is highly diverse and can be considered from multiple perspectives.

\subsection{Metrics}
\label{sec:metrics}

We establish three metrics to investigate the performance of different routers \( r_\theta \):
\underline{(1) \textbf{Original metric}} $\mu_o(r_\theta)$, i.e., the overall performance of the LLMs selected by router \( r_\theta \) on the given benchmark.
\underline{(2) \textbf{Reference value}} \( V_R \).  For each benchmark, we select a representative LLM with strong performance as the reference, such as GPT-4, with its performance denoted as \text{Perf.(ref.)}. Then,

\begin{equation}
    V_{R} = \mu_o(r_\theta)/\text{Perf.(ref.)}.
\end{equation}
See details of reference LLMs in Appendix \ref{sec:detai_routereval}.
\underline{(3) \textbf{Best single model value}} \( V_B \). In addition to the global metric \( V_G \), we also need a local metric to explore the potential of the router. Specifically, given a set of candidate models, let the best performance of a single model in the candidate set be \( \text{Perf.(BSM)} \). Then,
\begin{equation}
    V_B = \mu_o(r_\theta)/\text{Perf.(BSM)}.
\end{equation}
\underline{(4) \textbf{Classification bias}} \( E_p \). 
As mentioned in Section \ref{sec:extra}, under the current settings, the classification of the router is prone to bias if an effective training method for the router is not proposed. Therefore, we use entropy to measure the diversity of the classifier's prediction distribution. If the router always select same LLM, the entropy of its prediction distribution will be low, indicating a lack of diversity. Specifically,
\begin{equation}
    E_p = -\frac{1}{n}\sum\nolimits_{j=1}^n\sum\nolimits_{i=1}^m P_i^{(j)}\log P_i^{(j)},
\end{equation}
where \(P_i^{(j)} \in [0,1]^m\) represents the probability that the router selects the LLM with index \(i\) for the \(j\)-th test sample.

\subsection{Baselines}
\label{sec:baseline}

In this paper, we consider two kind of baselines for RouterEval. Specifically, these include:
\underline{(1) \textbf{Strong router}}. To directly evaluate the performance of routers, we introduce the oracle router \( r_o \) and \( r_o(0.5) \) mentioned in Section \ref{sec:poten} .
\underline{(2) \textbf{Existing router}}. We also compile some recently router methods, such as LinearR, MLPR, C-RoBERTa, MLC, and PRknn~\cite{hu2024routerbench,srivatsa2024harnessing,zhao2024model}. See Appendix \ref{sec:routers} or our code for detailed implementation.

\subsection{Result}

In this section, we present the results of the baselines shown in Section \ref{sec:baseline}, evaluated using the metrics described in Section \ref{sec:metrics}. The results under easy level settings, $m\in\{3,5\}$, are shown in Table \ref{tab:res1} and Table \ref{tab:res2}. See  more results under hard level settings $m \in \{10,100,1000\}$ in Appendix \ref{sec:res_hard}.

The experimental results show that most existing routers have some classification capability. However, the selected LLMs still lag significantly behind the best single models in most settings and the strong reference model in terms of performance, i.e., $V_R \leq 1$ and $V_B \leq 1$. Moreover, no single router consistently outperforms other router across different benchmarks. Additionally, some routers exhibit lower  $E_p$ values, where low entropy 
$E_p$ suggests potential overfitting, resulting in biases when selecting LLMs. For a detailed analysis, please refer to Section \ref{sec:ana}.

\begin{table*}[t]
  \centering

  \resizebox*{0.99\linewidth}{!}{
    \begin{tabular}{|c|l|rrrr|rrrr|rrrr|rrrr|}
    \toprule
    \multicolumn{1}{|r}{} &       & \multicolumn{4}{c|}{GPQA}      & \multicolumn{4}{c|}{MUSR} & \multicolumn{4}{c|}{MATH Lvl 5}     & \multicolumn{4}{c|}{MMLU-PRO} \\
\cmidrule{3-18}    \multicolumn{1}{|r}{} & Router & \multicolumn{1}{l}{$\mu_o$↑} & \multicolumn{1}{l}{$V_R$↑} & \multicolumn{1}{l}{$V_B$↑} & \multicolumn{1}{l|}{$E_p$} & \multicolumn{1}{l}{$\mu_o$↑} & \multicolumn{1}{l}{$V_R$↑} & \multicolumn{1}{l}{$V_B$↑} & \multicolumn{1}{l|}{$E_p$} & \multicolumn{1}{l}{$\mu_o$↑} & \multicolumn{1}{l}{$V_R$↑} & \multicolumn{1}{l}{$V_B$↑} & \multicolumn{1}{l|}{$E_p$} & \multicolumn{1}{l}{$\mu_o$↑} & \multicolumn{1}{l}{$V_R$↑} & \multicolumn{1}{l}{$V_B$↑} & \multicolumn{1}{l|}{$E_p$} \\
       \midrule
    \multirow{8}[1]{*}{\begin{sideways}$m=3$\end{sideways}} & \cellcolor[rgb]{ .996,  .89,  .871}Oracle $r_o$ & \cellcolor[rgb]{ .996,  .89,  .871}0.68  & \cellcolor[rgb]{ .996,  .89,  .871}1.71  & \cellcolor[rgb]{ .996,  .89,  .871}2.00  & \cellcolor[rgb]{ .996,  .89,  .871}0.72  & \cellcolor[rgb]{ .996,  .89,  .871}0.74  & \cellcolor[rgb]{ .996,  .89,  .871}1.05  & \cellcolor[rgb]{ .996,  .89,  .871}1.74  & \cellcolor[rgb]{ .996,  .89,  .871}0.81  & \cellcolor[rgb]{ .996,  .89,  .871}0.56  & \cellcolor[rgb]{ .996,  .89,  .871}1.39  & \cellcolor[rgb]{ .996,  .89,  .871}1.33  & \cellcolor[rgb]{ .996,  .89,  .871}1.14  & \cellcolor[rgb]{ .996,  .89,  .871}0.67  & \cellcolor[rgb]{ .996,  .89,  .871}1.05  & \cellcolor[rgb]{ .996,  .89,  .871}1.45  & \cellcolor[rgb]{ .996,  .89,  .871}1.05  \\
          & \cellcolor[rgb]{ .996,  .89,  .871}$r_o(0.5)$ & \cellcolor[rgb]{ .996,  .89,  .871}0.49  & \cellcolor[rgb]{ .996,  .89,  .871}1.24  & \cellcolor[rgb]{ .996,  .89,  .871}1.43  & \cellcolor[rgb]{ .996,  .89,  .871}1.41  & \cellcolor[rgb]{ .996,  .89,  .871}0.56  & \cellcolor[rgb]{ .996,  .89,  .871}0.81  & \cellcolor[rgb]{ .996,  .89,  .871}1.32  & \cellcolor[rgb]{ .996,  .89,  .871}1.43  & \cellcolor[rgb]{ .996,  .89,  .871}0.45  & \cellcolor[rgb]{ .996,  .89,  .871}1.14  & \cellcolor[rgb]{ .996,  .89,  .871}1.06  & \cellcolor[rgb]{ .996,  .89,  .871}1.49  & \cellcolor[rgb]{ .996,  .89,  .871}0.55  & \cellcolor[rgb]{ .996,  .89,  .871}0.86  & \cellcolor[rgb]{ .996,  .89,  .871}1.13  & \cellcolor[rgb]{ .996,  .89,  .871}1.48  \\
          & \cellcolor[rgb]{ .854, .909, .988}LinearR & \cellcolor[rgb]{ .854, .909, .988}\underline{\textbf{0.39}} & \cellcolor[rgb]{ .854, .909, .988}\underline{\textbf{0.99}} & \cellcolor[rgb]{ .854, .909, .988}\underline{\textbf{1.14}} & \cellcolor[rgb]{ .854, .909, .988}1.48  & \cellcolor[rgb]{ .854, .909, .988}0.43  & \cellcolor[rgb]{ .854, .909, .988}0.61  & \cellcolor[rgb]{ .854, .909, .988}0.99  & \cellcolor[rgb]{ .854, .909, .988}1.43  & \cellcolor[rgb]{ .854, .909, .988}0.44  & \cellcolor[rgb]{ .854, .909, .988}1.11  & \cellcolor[rgb]{ .854, .909, .988}0.95  & \cellcolor[rgb]{ .854, .909, .988}1.37  & \cellcolor[rgb]{ .854, .909, .988}\underline{\textbf{0.54}}  & \cellcolor[rgb]{ .854, .909, .988}\underline{\textbf{0.85}}  & \cellcolor[rgb]{ .854, .909, .988}\underline{\textbf{1.01}}  & \cellcolor[rgb]{ .854, .909, .988}1.37  \\
          & \cellcolor[rgb]{ .854, .909, .988}MLPR & \cellcolor[rgb]{ .854, .909, .988}0.36  & \cellcolor[rgb]{ .854, .909, .988}0.91  & \cellcolor[rgb]{ .854, .909, .988}1.05  & \cellcolor[rgb]{ .854, .909, .988}1.29  & \cellcolor[rgb]{ .854, .909, .988}0.43  & \cellcolor[rgb]{ .854, .909, .988}0.62  & \cellcolor[rgb]{ .854, .909, .988}0.99  & \cellcolor[rgb]{ .854, .909, .988}1.40  & \cellcolor[rgb]{ .854, .909, .988}0.43  & \cellcolor[rgb]{ .854, .909, .988}1.08  & \cellcolor[rgb]{ .854, .909, .988}0.92  & \cellcolor[rgb]{ .854, .909, .988}1.32  & \cellcolor[rgb]{ .854, .909, .988}\underline{\textbf{0.54}} & \cellcolor[rgb]{ .854, .909, .988}\underline{\textbf{0.85}} & \cellcolor[rgb]{ .854, .909, .988}\underline{\textbf{1.01}} & \cellcolor[rgb]{ .854, .909, .988}1.36  \\
          & \cellcolor[rgb]{ .854, .909, .988}C-RoBERTa & \cellcolor[rgb]{ .854, .909, .988}0.33  & \cellcolor[rgb]{ .854, .909, .988}0.82  & \cellcolor[rgb]{ .854, .909, .988}0.92  & \cellcolor[rgb]{ .854, .909, .988}0.60  & \cellcolor[rgb]{ .854, .909, .988}\underline{\textbf{0.44}} & \cellcolor[rgb]{ .854, .909, .988}\underline{\textbf{0.63}} & \cellcolor[rgb]{ .854, .909, .988}\underline{\textbf{1.01}} & \cellcolor[rgb]{ .854, .909, .988}1.02  & \cellcolor[rgb]{ .854, .909, .988}\underline{\textbf{0.45}} & \cellcolor[rgb]{ .854, .909, .988}\underline{\textbf{1.13}} & \cellcolor[rgb]{ .854, .909, .988}\underline{\textbf{0.99}} & \cellcolor[rgb]{ .854, .909, .988}0.65  & \cellcolor[rgb]{ .854, .909, .988}0.54  & \cellcolor[rgb]{ .854, .909, .988}0.84  & \cellcolor[rgb]{ .854, .909, .988}0.97  & \cellcolor[rgb]{ .854, .909, .988}0.33  \\
          & \cellcolor[rgb]{ .854, .909, .988}MLC & \cellcolor[rgb]{ .854, .909, .988}0.30  & \cellcolor[rgb]{ .854, .909, .988}0.75  & \cellcolor[rgb]{ .854, .909, .988}0.93  & \cellcolor[rgb]{ .854, .909, .988}0.00  & \cellcolor[rgb]{ .854, .909, .988}0.39  & \cellcolor[rgb]{ .854, .909, .988}0.56  & \cellcolor[rgb]{ .854, .909, .988}0.91  & \cellcolor[rgb]{ .854, .909, .988}0.00  & \cellcolor[rgb]{ .854, .909, .988}0.42  & \cellcolor[rgb]{ .854, .909, .988}1.06  & \cellcolor[rgb]{ .854, .909, .988}0.88  & \cellcolor[rgb]{ .854, .909, .988}0.53  & \cellcolor[rgb]{ .854, .909, .988}0.53  & \cellcolor[rgb]{ .854, .909, .988}0.84  & \cellcolor[rgb]{ .854, .909, .988}0.95  & \cellcolor[rgb]{ .854, .909, .988}0.49  \\
          & \cellcolor[rgb]{ .854, .909, .988}PRknn & \cellcolor[rgb]{ .854, .909, .988}0.32  & \cellcolor[rgb]{ .854, .909, .988}0.81  & \cellcolor[rgb]{ .854, .909, .988}0.93  & \cellcolor[rgb]{ .854, .909, .988}1.56  & \cellcolor[rgb]{ .854, .909, .988}0.40  & \cellcolor[rgb]{ .854, .909, .988}0.57  & \cellcolor[rgb]{ .854, .909, .988}0.92  & \cellcolor[rgb]{ .854, .909, .988}1.56  & \cellcolor[rgb]{ .854, .909, .988}0.40  & \cellcolor[rgb]{ .854, .909, .988}1.01  & \cellcolor[rgb]{ .854, .909, .988}0.88  & \cellcolor[rgb]{ .854, .909, .988}1.56  & \cellcolor[rgb]{ .854, .909, .988}0.50  & \cellcolor[rgb]{ .854, .909, .988}0.79  & \cellcolor[rgb]{ .854, .909, .988}0.92  & \cellcolor[rgb]{ .854, .909, .988}1.56  \\
          & \cellcolor[rgb]{ .854, .909, .988}Random & \cellcolor[rgb]{ .854, .909, .988}0.30  & \cellcolor[rgb]{ .854, .909, .988}0.76  & \cellcolor[rgb]{ .854, .909, .988}0.87  & \cellcolor[rgb]{ .854, .909, .988}1.59  & \cellcolor[rgb]{ .854, .909, .988}0.39  & \cellcolor[rgb]{ .854, .909, .988}0.56  & \cellcolor[rgb]{ .854, .909, .988}0.90  & \cellcolor[rgb]{ .854, .909, .988}1.59  & \cellcolor[rgb]{ .854, .909, .988}0.35  & \cellcolor[rgb]{ .854, .909, .988}0.88  & \cellcolor[rgb]{ .854, .909, .988}0.79  & \cellcolor[rgb]{ .854, .909, .988}1.59  & \cellcolor[rgb]{ .854, .909, .988}0.43  & \cellcolor[rgb]{ .854, .909, .988}0.67  & \cellcolor[rgb]{ .854, .909, .988}0.82  & \cellcolor[rgb]{ .854, .909, .988}1.59  \\
           \midrule
    \multirow{8}[0]{*}{\begin{sideways}$m=5$\end{sideways}} & \cellcolor[rgb]{ .996,  .89,  .871}Oracle $r_o$ & \cellcolor[rgb]{ .996,  .89,  .871}0.87  & \cellcolor[rgb]{ .996,  .89,  .871}2.18  & \cellcolor[rgb]{ .996,  .89,  .871}2.52  & \cellcolor[rgb]{ .996,  .89,  .871}0.84  & \cellcolor[rgb]{ .996,  .89,  .871}0.82  & \cellcolor[rgb]{ .996,  .89,  .871}1.17  & \cellcolor[rgb]{ .996,  .89,  .871}1.77  & \cellcolor[rgb]{ .996,  .89,  .871}1.31  & \cellcolor[rgb]{ .996,  .89,  .871}0.61  & \cellcolor[rgb]{ .996,  .89,  .871}1.51  & \cellcolor[rgb]{ .996,  .89,  .871}1.52  & \cellcolor[rgb]{ .996,  .89,  .871}1.60  & \cellcolor[rgb]{ .996,  .89,  .871}0.72  & \cellcolor[rgb]{ .996,  .89,  .871}1.14  & \cellcolor[rgb]{ .996,  .89,  .871}1.69  & \cellcolor[rgb]{ .996,  .89,  .871}1.50  \\
          & \cellcolor[rgb]{ .996,  .89,  .871}$r_o(0.5)$ & \cellcolor[rgb]{ .996,  .89,  .871}0.58  & \cellcolor[rgb]{ .996,  .89,  .871}1.47  & \cellcolor[rgb]{ .996,  .89,  .871}1.69  & \cellcolor[rgb]{ .996,  .89,  .871}1.99  & \cellcolor[rgb]{ .996,  .89,  .871}0.61  & \cellcolor[rgb]{ .996,  .89,  .871}0.87  & \cellcolor[rgb]{ .996,  .89,  .871}1.30  & \cellcolor[rgb]{ .996,  .89,  .871}2.10  & \cellcolor[rgb]{ .996,  .89,  .871}0.46  & \cellcolor[rgb]{ .996,  .89,  .871}1.16  & \cellcolor[rgb]{ .996,  .89,  .871}1.13  & \cellcolor[rgb]{ .996,  .89,  .871}2.16  & \cellcolor[rgb]{ .996,  .89,  .871}0.56  & \cellcolor[rgb]{ .996,  .89,  .871}0.88  & \cellcolor[rgb]{ .996,  .89,  .871}1.23  & \cellcolor[rgb]{ .996,  .89,  .871}2.14  \\
          & \cellcolor[rgb]{ .854, .909, .988}LinearR & \cellcolor[rgb]{ .854, .909, .988}\underline{\textbf{0.40}} & \cellcolor[rgb]{ .854, .909, .988}\underline{\textbf{1.00}} & \cellcolor[rgb]{ .854, .909, .988}\underline{\textbf{1.15}} & \cellcolor[rgb]{ .854, .909, .988}2.19  & \cellcolor[rgb]{ .854, .909, .988}\underline{\textbf{0.46}} & \cellcolor[rgb]{ .854, .909, .988}\underline{\textbf{0.65}} & \cellcolor[rgb]{ .854, .909, .988}\underline{\textbf{0.96}} & \cellcolor[rgb]{ .854, .909, .988}2.24  & \cellcolor[rgb]{ .854, .909, .988}0.43  & \cellcolor[rgb]{ .854, .909, .988}1.08  & \cellcolor[rgb]{ .854, .909, .988}0.97  & \cellcolor[rgb]{ .854, .909, .988}2.12  & \cellcolor[rgb]{ .854, .909, .988}0.54  & \cellcolor[rgb]{ .854, .909, .988}0.85  & \cellcolor[rgb]{ .854, .909, .988}1.01  & \cellcolor[rgb]{ .854, .909, .988}2.06  \\
          & \cellcolor[rgb]{ .854, .909, .988}MLPR & \cellcolor[rgb]{ .854, .909, .988}0.36  & \cellcolor[rgb]{ .854, .909, .988}0.92  & \cellcolor[rgb]{ .854, .909, .988}1.05  & \cellcolor[rgb]{ .854, .909, .988}2.04  & \cellcolor[rgb]{ .854, .909, .988}0.45  & \cellcolor[rgb]{ .854, .909, .988}0.65  & \cellcolor[rgb]{ .854, .909, .988}0.95  & \cellcolor[rgb]{ .854, .909, .988}2.24  & \cellcolor[rgb]{ .854, .909, .988}\underline{\textbf{0.44}} & \cellcolor[rgb]{ .854, .909, .988}\underline{\textbf{1.10}} & \cellcolor[rgb]{ .854, .909, .988}\underline{\textbf{0.99}} & \cellcolor[rgb]{ .854, .909, .988}2.06  & \cellcolor[rgb]{ .854, .909, .988}\underline{\textbf{0.54}} & \cellcolor[rgb]{ .854, .909, .988}\underline{\textbf{0.86}} & \cellcolor[rgb]{ .854, .909, .988}\underline{\textbf{1.02}} & \cellcolor[rgb]{ .854, .909, .988}2.05  \\
          & \cellcolor[rgb]{ .854, .909, .988}C-RoBERTa & \cellcolor[rgb]{ .854, .909, .988}0.33  & \cellcolor[rgb]{ .854, .909, .988}0.84  & \cellcolor[rgb]{ .854, .909, .988}0.95  & \cellcolor[rgb]{ .854, .909, .988}0.82  & \cellcolor[rgb]{ .854, .909, .988}0.46  & \cellcolor[rgb]{ .854, .909, .988}0.65  & \cellcolor[rgb]{ .854, .909, .988}0.93  & \cellcolor[rgb]{ .854, .909, .988}1.54  & \cellcolor[rgb]{ .854, .909, .988}0.42  & \cellcolor[rgb]{ .854, .909, .988}1.05  & \cellcolor[rgb]{ .854, .909, .988}0.91  & \cellcolor[rgb]{ .854, .909, .988}0.65  & \cellcolor[rgb]{ .854, .909, .988}0.53  & \cellcolor[rgb]{ .854, .909, .988}0.84  & \cellcolor[rgb]{ .854, .909, .988}0.97  & \cellcolor[rgb]{ .854, .909, .988}0.33  \\
          & \cellcolor[rgb]{ .854, .909, .988}MLC & \cellcolor[rgb]{ .854, .909, .988}0.30  & \cellcolor[rgb]{ .854, .909, .988}0.76  & \cellcolor[rgb]{ .854, .909, .988}0.86  & \cellcolor[rgb]{ .854, .909, .988}0.00  & \cellcolor[rgb]{ .854, .909, .988}0.40  & \cellcolor[rgb]{ .854, .909, .988}0.58  & \cellcolor[rgb]{ .854, .909, .988}0.85  & \cellcolor[rgb]{ .854, .909, .988}0.09  & \cellcolor[rgb]{ .854, .909, .988}0.41  & \cellcolor[rgb]{ .854, .909, .988}1.04  & \cellcolor[rgb]{ .854, .909, .988}0.87  & \cellcolor[rgb]{ .854, .909, .988}0.77  & \cellcolor[rgb]{ .854, .909, .988}0.54  & \cellcolor[rgb]{ .854, .909, .988}0.85  & \cellcolor[rgb]{ .854, .909, .988}1.00  & \cellcolor[rgb]{ .854, .909, .988}0.74  \\
          & \cellcolor[rgb]{ .854, .909, .988}PRknn & \cellcolor[rgb]{ .854, .909, .988}0.32  & \cellcolor[rgb]{ .854, .909, .988}0.80  & \cellcolor[rgb]{ .854, .909, .988}0.91  & \cellcolor[rgb]{ .854, .909, .988}2.30  & \cellcolor[rgb]{ .854, .909, .988}0.43  & \cellcolor[rgb]{ .854, .909, .988}0.61  & \cellcolor[rgb]{ .854, .909, .988}0.90  & \cellcolor[rgb]{ .854, .909, .988}2.30  & \cellcolor[rgb]{ .854, .909, .988}0.39  & \cellcolor[rgb]{ .854, .909, .988}0.97  & \cellcolor[rgb]{ .854, .909, .988}0.88  & \cellcolor[rgb]{ .854, .909, .988}2.30  & \cellcolor[rgb]{ .854, .909, .988}0.49  & \cellcolor[rgb]{ .854, .909, .988}0.76  & \cellcolor[rgb]{ .854, .909, .988}0.93  & \cellcolor[rgb]{ .854, .909, .988}2.30  \\
          & \cellcolor[rgb]{ .854, .909, .988}Random & \cellcolor[rgb]{ .854, .909, .988}0.30  & \cellcolor[rgb]{ .854, .909, .988}0.75  & \cellcolor[rgb]{ .854, .909, .988}0.85  & \cellcolor[rgb]{ .854, .909, .988}2.32  & \cellcolor[rgb]{ .854, .909, .988}0.40  & \cellcolor[rgb]{ .854, .909, .988}0.57  & \cellcolor[rgb]{ .854, .909, .988}0.84  & \cellcolor[rgb]{ .854, .909, .988}2.32  & \cellcolor[rgb]{ .854, .909, .988}0.32  & \cellcolor[rgb]{ .854, .909, .988}0.80  & \cellcolor[rgb]{ .854, .909, .988}0.73  & \cellcolor[rgb]{ .854, .909, .988}2.32  & \cellcolor[rgb]{ .854, .909, .988}0.40  & \cellcolor[rgb]{ .854, .909, .988}0.62  & \cellcolor[rgb]{ .854, .909, .988}0.78  & \cellcolor[rgb]{ .854, .909, .988}2.32  \\
           \midrule
    \end{tabular}%
    }
      \caption{\textbf{The Results on RouterEval (part2)}. See Section \ref{sec:metrics} for detials of various metrics.  Red area and blue area highlights indicate the "Strong router" and "Existing router" mentioned in Section \ref{sec:baseline}, respectively. The best results in existing router methods are highlighted with underlines and on bold. The values in the table are rounded to two decimal places. The results on hard level settings are shown in Appendix \ref{sec:res_hard}.}
  \label{tab:res2}%
\end{table*}%

\section{Analysis}
\label{sec:ana}
In this Section, We conduct a more comprehensive analysis of the proposed RouterEval.

\begin{mdframed}[backgroundcolor=gray!8]
\begin{minipage}{\linewidth}
\vspace{2pt}
\noindent (1) The differences between Routing LLM and existing paradigms.
\vspace{2pt}
\end{minipage}
\end{mdframed}
In fact, several existing paradigms in other fields are related to Routing LLMs. This section aims to analyze their relationships and differences. See related works in Appendix \ref{sec:related} for more details.
\begin{figure}[t]
  \includegraphics[width=0.95\linewidth]{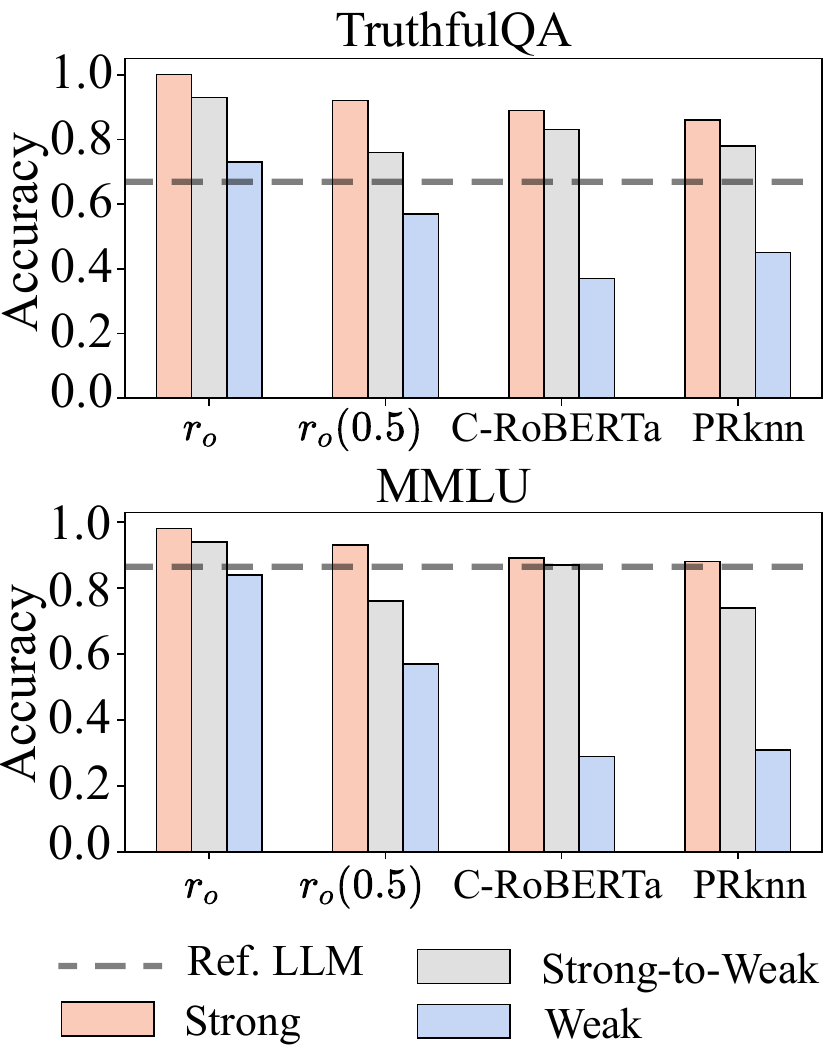}
  \caption{\textbf{The Results on Different Candidate Group}. }
  \vspace{-15pt}
  \label{fig:group}
\end{figure}

\noindent\underline{\textbf{Recommender system.}}
In fact, Routing LLMs are a specialized type of Recommender System (RS), where the input can be seen as the user in the RS, LLM candidates as the items to be recommended, and the performance record as the interaction history between items and users. Given the input, the router needs to recommend an appropriate LLM to achieve various objectives, such as high accuracy, low computational cost, or minimal hallucination. However, compared to traditional RS, Routing LLMs have very limited “user information” that can be collected, and labeling this data is challenging. Most of the data is private and rarely open-source. The over 200,000,000 performance records we have organized and open-sourced represent only a small step toward building a router with strong recommendation capabilities.

\noindent\underline{\textbf{LLMs ensemble.}} 
For a given input, the LLMs ensemble paradigm focuses more on requiring all LLM candidates to perform inference, then aggregating and organizing their results, with the final output achieved through strategies such as majority voting. In contrast, the Routing LLMs paradigm performs an efficient assignment before the candidates’ inference, making it a more computationally efficient approach. Of course, since different paradigms can borrow from each other, there is still some technical overlap between these two paradigms in many improvement methods.

\noindent\underline{\textbf{LLMs fusion.}} 
The paradigm of LLM fusion, which combines different LLMs to achieve superior performance, is a promising technology. It typically requires the LLMs being fused to have the same structure. However, Routing LLMs can integrate not only homogenous models but also heterogeneous ones, offering greater flexibility from an application perspective.

\noindent\underline{\textbf{Mixture-of-Experts (MoE).}}  
Traditional MoE mainly focuses on the mixture of local parameters, such as FFN, within a single LLM to achieve better performance like a larger model by activating only a subset of parameters. Similarly, Routing LLMs can be viewed as a larger-granularity MoE, specifically a Model-level “MoE”, where the experts refer to the candidate LLMs. Both are effective methods for improving LLM performance.

In summary, although the Routing LLMs paradigm shares similarities with many existing approaches, it is important to note that Routing LLMs can actually be compatible with these paradigms. This compatibility allows the integration of heterogeneous models from these paradigms into the candidate pool, leading to further enhancements in performance. 

\begin{mdframed}[backgroundcolor=gray!8]
\begin{minipage}{\linewidth}
\vspace{2pt}
\noindent (2) How do different types of candidate combinations affect performance?
\vspace{2pt}
\end{minipage}
\end{mdframed}
In Section \ref{sec:constr}, we constructed three types of candidate combinations, namely "all-strong," "all-weak," and "strong-to-weak" groups. In this part, we attempt to analyze how the selection of these different groups affects the performance of Routing LLMs. As shown in Fig.~\ref{fig:group}, taking $m=5$ as example, we consider two types of strong routers, $r_o$ and $r_o(0.5)$, as well as two existing routers, C-RoBERTa and PRknn. 

We observe that, overall, the "all-strong" group typically exhibits the best performance, while the "all-weak" group lags behind. However, when the router has sufficiently strong classification capabilities, like $r_o$, even the "all-weak" group can achieve satisfactory performance. For instance, in MMLU result in Fig.~\ref{fig:group}, the performance of each individual LLM in the "all-weak" group does not exceed 0.3, but under the capable router $r_o$, it can approximatively approach the performance of reference model, i.e., GPT-4. This indicates that heterogeneous models can effectively complement each other under the guidance of Routing LLMs, provided that the router is sufficiently capable.

\begin{table}[htbp]
  \centering

  \resizebox*{0.99\linewidth}{!}{
    \begin{tabular}{|c|c|ccc|}
    \toprule
          & \multicolumn{1}{c|}{Router} & \multicolumn{1}{l}{all-strong} & \multicolumn{1}{l}{all-weak} & \multicolumn{1}{l|}{strong-to-weak} \\
    \midrule
    \multirow{8}[4]{*}{\begin{sideways}$m=3$\end{sideways}} & \cellcolor[rgb]{ .996,  .89,  .871}Oracle $r_o$ & \cellcolor[rgb]{ .996,  .89,  .871}1.39  & \cellcolor[rgb]{ .996,  .89,  .871}0.77  & \cellcolor[rgb]{ .996,  .89,  .871}0.96  \\
          & \cellcolor[rgb]{ .996,  .89,  .871}$r_o(0.5)$ & \cellcolor[rgb]{ .996,  .89,  .871}1.55  & \cellcolor[rgb]{ .996,  .89,  .871}1.42  & \cellcolor[rgb]{ .996,  .89,  .871}1.45  \\
\cmidrule{2-5}          & \cellcolor[rgb]{ .854, .909, .988}LinearR & \cellcolor[rgb]{ .854, .909, .988}1.54  & \cellcolor[rgb]{ .854, .909, .988}1.54  & \cellcolor[rgb]{ .854, .909, .988}0.81  \\
          & \cellcolor[rgb]{ .854, .909, .988}MLPR & \cellcolor[rgb]{ .854, .909, .988}1.50  & \cellcolor[rgb]{ .854, .909, .988}1.52  & \cellcolor[rgb]{ .854, .909, .988}0.76  \\
          & \cellcolor[rgb]{ .854, .909, .988}C-RoBERTa & \cellcolor[rgb]{ .854, .909, .988}0.93  & \cellcolor[rgb]{ .854, .909, .988}0.94  & \cellcolor[rgb]{ .854, .909, .988}0.00  \\
          & \cellcolor[rgb]{ .854, .909, .988}MLC & \cellcolor[rgb]{ .854, .909, .988}1.52  & \cellcolor[rgb]{ .854, .909, .988}0.34  & \cellcolor[rgb]{ .854, .909, .988}0.52  \\
          & \cellcolor[rgb]{ .854, .909, .988}PRknn & \cellcolor[rgb]{ .854, .909, .988}1.58  & \cellcolor[rgb]{ .854, .909, .988}1.56  & \cellcolor[rgb]{ .854, .909, .988}1.52  \\
          & \cellcolor[rgb]{ .854, .909, .988}Random & \cellcolor[rgb]{ .854, .909, .988}1.59  & \cellcolor[rgb]{ .854, .909, .988}1.59  & \cellcolor[rgb]{ .854, .909, .988}1.59  \\
    \midrule
    \multirow{8}[4]{*}{\begin{sideways}$m=5$\end{sideways}} & \cellcolor[rgb]{ .996,  .89,  .871}Oracle $r_o$ & \cellcolor[rgb]{ .996,  .89,  .871}2.09  & \cellcolor[rgb]{ .996,  .89,  .871}0.90  & \cellcolor[rgb]{ .996,  .89,  .871}1.49  \\
          & \cellcolor[rgb]{ .996,  .89,  .871}$r_o(0.5)$ & \cellcolor[rgb]{ .996,  .89,  .871}2.27  & \cellcolor[rgb]{ .996,  .89,  .871}2.00  & \cellcolor[rgb]{ .996,  .89,  .871}2.15  \\
\cmidrule{2-5}          & \cellcolor[rgb]{ .854, .909, .988}LinearR & \cellcolor[rgb]{ .854, .909, .988}2.27  & \cellcolor[rgb]{ .854, .909, .988}2.28  & \cellcolor[rgb]{ .854, .909, .988}1.58  \\
          & \cellcolor[rgb]{ .854, .909, .988}MLPR & \cellcolor[rgb]{ .854, .909, .988}2.26  & \cellcolor[rgb]{ .854, .909, .988}2.25  & \cellcolor[rgb]{ .854, .909, .988}1.50  \\
          & \cellcolor[rgb]{ .854, .909, .988}C-RoBERTa & \cellcolor[rgb]{ .854, .909, .988}1.53  & \cellcolor[rgb]{ .854, .909, .988}1.53  & \cellcolor[rgb]{ .854, .909, .988}0.00  \\
          & \cellcolor[rgb]{ .854, .909, .988}MLC & \cellcolor[rgb]{ .854, .909, .988}2.25  & \cellcolor[rgb]{ .854, .909, .988}0.03  & \cellcolor[rgb]{ .854, .909, .988}1.06  \\
          & \cellcolor[rgb]{ .854, .909, .988}PRknn & \cellcolor[rgb]{ .854, .909, .988}2.31  & \cellcolor[rgb]{ .854, .909, .988}2.30  & \cellcolor[rgb]{ .854, .909, .988}2.28  \\
          & \cellcolor[rgb]{ .854, .909, .988}Random & \cellcolor[rgb]{ .854, .909, .988}2.32  & \cellcolor[rgb]{ .854, .909, .988}2.32  & \cellcolor[rgb]{ .854, .909, .988}2.32  \\
    \bottomrule
    \end{tabular}%
    }
      \caption{\textbf{The $E_p$ on Various Candidate Groups}. Some router methods suffer from the classification bias, i.e., $E_p$ is close to 0.}
  \label{tab:ep_group}%
\end{table}%

\begin{mdframed}[backgroundcolor=gray!8]
\begin{minipage}{\linewidth}
\vspace{2pt}
\noindent (3) Classification bias in routers.
\vspace{2pt}
\end{minipage}
\end{mdframed}
In Section \ref{sec:metrics}, we proposed using $E_p$ to analyze the classification bias of routers and found that existing router methods may exhibit classification bias, as shown in Tables \ref{tab:res1} and \ref{tab:res2}. In this part, taking MMLU benchmark as an example, we conduct a more detailed analysis. As illustrated in Table \ref{tab:ep_group}, the bias is  severe in several settings and router methods, where strong models are more likely to be selected with higher probability. 

In fact, if $E_p \to 0$
, the router degrades into an individual router, which is actually the best-performing router in the training set. While this strategy can still achieve decent performance, it fails to leverage the advantages of the Routing LLMs paradigm. From the results in Table \ref{tab:ep_group}, even in the "strong-to-weak" group, a strong router like $r_o$
 requires diverse selections to integrate the complementary strengths of different candidates for better performance. Thus, debiasing is crucial for further enhancing the router's capability.

\begin{mdframed}[backgroundcolor=gray!8]
\begin{minipage}{\linewidth}
\vspace{2pt}
\noindent (4) How to boost the performance of router?
\vspace{2pt}
\end{minipage}
\end{mdframed}
As shown in Table \ref{tab:res1} and \ref{tab:res2}, the current router has significant room for improvement, with potential directions including the following. First, fully leveraging the extra data of performance records in Section \ref{sec:extra} may be crucial. For instance, data augmentation, pre-training methods, and few-shot learning techniques could be constructed based on this data. Additionally, as mentioned in Section \ref{sec:ana} (1), if we regard Routing LLMs as a recommender system (RS), future research could focus on utilizing extra data to design effective representation learning for LLMs and inputs~\cite{zhuang2024embedllm}, addressing the cold start~\cite{schein2002methods,lam2008addressing,wei2021contrastive} problem of the router, and employing causal inference techniques~\cite{liang2016causal,wang2020causal,huang2025causality}, which are classic methods in RS research, to improve the router, especially for debiasing. This aligns with the objectives outlined in Section \ref{sec:ana} (3).

\begin{mdframed}[backgroundcolor=gray!8]
\begin{minipage}{\linewidth}
\vspace{2pt}
\noindent (5) Why not consider other routing objective, like computational cost, temporarily?
\vspace{2pt}
\end{minipage}
\end{mdframed}
In fact, the paradigm of RouterEval can be extended to consider additional objectives~\cite{chen2023frugalgpt,wang2025mixllm,vsakota2024fly} such as computational cost and hallucination rate through multi-objective optimization approaches~\cite{deb2005searching,giagkiozis2015methods}. However, as shown in the experiments in Section~\ref{sec:exp}, even when focusing solely on performance metrics across different benchmarks, the current router methods still have significant room for improvement. Under these circumstances, it is advisable to temporarily defer the exploration of other objectives. Otherwise, with the current limited data availability, adding more learning targets may further compromise performance. Therefore, it is recommended to first concentrate on enhancing performance. Once a certain level of exploration has been achieved, more data can then be utilized to expand to other optimization objectives.

\section{Conclusion}

In this paper, we comprehensively explored the potential of the Routing LLMs paradigm through extensive experiments and identified the scaling-up phenomenon of LLMs at the model level. Furthermore, given that the development of routers is hindered by the lack of comprehensive benchmarks, we introduce RouterEval, a benchmark based on  200 million performance records across 12 LLM  evaluations. The evaluations reveal that existing routing methods still have room for improvement.

\clearpage

\section*{Limitations}

Although we observed that the Routing LLMs paradigm can trigger the model-level scaling-up phenomenon of LLMs in this paper, this implies that a large number of LLMs may cause deployment challenges. Fortunately, our experiments found that the cost-effectiveness of this paradigm is highest when there are approximately 3 to 10 LLM candidates, that is, the performance growth rate is the fastest with fewer LLMs. Therefore, if we do not pursue extremely outstanding performance, it is still possible to achieve lower computational requirements in deployment. Additionally, for industrial deployment, when there are sufficiently many inputs as a batch of requests, in fact, as long as the routing infrastructure is well-developed, the average computational cost of inputs is not high. Furthermore, despite considering a sufficient number of LLMs and their corresponding performance records, the current data volume still cannot produce an excellent router. Increasing the data volume will be one of the most important research directions in the future, as most benchmark performance record data are not open-source and really expensive, which requires the cooperation of the entire community.

\bibliography{custom}

\clearpage
\appendix

\section{The More Details of RouterEval}
\label{sec:detai_routereval}

In this paper, we have considered 12 benchmarks across areas such as knowledge-based Q\&A, commonsense reasoning, semantic understanding, mathematical reasoning, and instruction following, etc., including ARC, HellaSwag, MMLU, TruthfulQA, Winogrande, GSM8k, IFEval, BBH, GPQA, MUSR, MATH Lvl 5, and MMLU-PRO~\cite{zellers2019hellaswag,clark2018think,wang2024mmlu,hendrycks2020measuring,lin2021truthfulqa,sakaguchi2021winogrande,cobbe2021training,zhou2023instruction,suzgun2022challenging,rein2024gpqa,sprague2023musr}. Next, the test samples for each benchmark are split into training, validation, and test sets in a ratio of 8:1:1, as detailed in Table~\ref{tab:detail_routereval}.

Furthermore, as discussed in Section~\ref{sec:metrics}, we need to designate a representative and high-performing LLM for each benchmark to construct a reference value. For most benchmarks, GPT-4, given its widespread application and popularity, is highly suitable for constructing reference values. However, some benchmarks lack evaluation results for GPT-4. Therefore, we individually identify representative LLMs from these benchmarks to construct reference values. When the final performance of the model constructed under the Routing LLMs paradigm exceeds a reference value of 1, it indicates that the mechanism can enable relatively weak candidates to collaborate and surpass the performance of some well-known commercial LLMs.

\begin{table*}[t]
  \centering
 \resizebox*{0.99\linewidth}{!}{
    \begin{tabular}{lrrrrrr}
    \toprule
    \textbf{Benchmark} & \multicolumn{1}{l}{\textbf{\#LLMs}} & \multicolumn{1}{l}{\textbf{\#Train}} & \multicolumn{1}{l}{\textbf{\#Val}} & \multicolumn{1}{l}{\textbf{\#Test}} & \multicolumn{1}{l}{\textbf{Ref. Per.}} & \multicolumn{1}{l}{\textbf{Ref. Name}} \\
    \midrule
    ARC~\cite{clark2018think}   & 5000  & 937 & 117   & 118   & \multicolumn{1}{c}{0.852}  & \multicolumn{1}{l}{GPT-3.5} \\
    HellaSwag~\cite{zellers2019hellaswag} &   5000 & 8033    &   1004    &   1005    &  \multicolumn{1}{c}{0.953} &  \multicolumn{1}{l}{GPT-4} \\
    MMLU~\cite{wang2024mmlu}  &   5000    &   11215    &    1397   &   1430    &   \multicolumn{1}{c}{0.864}    & \multicolumn{1}{l}{GPT-4} \\
    TruthfulQA~\cite{lin2021truthfulqa} &   5000    &    653   &    82   &     82  &   \multicolumn{1}{c}{0.669}    & \multicolumn{1}{l}{Qwen1.5-32B} \\
    Winogrande~\cite{sakaguchi2021winogrande} &   5000    &   1013    & 127      &   127    &    \multicolumn{1}{c}{0.875}    & \multicolumn{1}{l}{GPT-4} \\
    GSM8k~\cite{cobbe2021training} & 5000      &    1055   &    132   &    132   &   \multicolumn{1}{c}{0.920}    &  \multicolumn{1}{l}{GPT-4} \\
    IFEval~\cite{zhou2023instruction} &    3811   &   432    &    54   &     55  &   \multicolumn{1}{c}{0.769}    &  \multicolumn{1}{l}{GPT-4} \\
    BBH~\cite{suzgun2022challenging}   &   3811    &   4607    &   577    &    577   &    \multicolumn{1}{c}{0.830}   & \multicolumn{1}{l}{GPT-4}  \\
    GPQA~\cite{rein2024gpqa}  &  3811     &    952   &  120     &   120    &    \multicolumn{1}{c}{0.397}   & \multicolumn{1}{l}{GPT-4} \\
    MUSR~\cite{sprague2023musr}  &   3811    &    604   &   76    &     76  &    \multicolumn{1}{c}{0.699}   &  \multicolumn{1}{l}{GPT-4} \\
    MATH Lvl 5~\cite{hendrycks2021measuring} &    3811   &   1057    &  131     &   136    &   \multicolumn{1}{c}{0.400}    &  \multicolumn{1}{l}{GPT-4} \\
    MMMU-PRO~\cite{wang2024mmlu} &    1823   &    9625   &  1203     &   1204    &    \multicolumn{1}{c}{0.637}   &  \multicolumn{1}{l}{GPT-4 Turbo} \\
    \bottomrule
    \end{tabular}%
    }
     \caption{\textbf{The Details of Each Benchmark in RouterEval.} The notation "\#LLMs" represents the number of LLMs included in the performance records of each benchmark. The terms "\#Train/\#Val/\#Test" denote the respective counts of training, validation, and test datasets constructed for each benchmark within RouterEval. "Ref.Name" and "Ref.Per." indicate the name and performance of the LLM used to establish the reference value in Section \ref{sec:metrics}. Avatar
All benchmarks involve a total of 8,576 LLMs. In RouterEval, we additionally provide over 200,000,000 performance records of these LLMs across different benchmarks.}
  \label{tab:detail_routereval}%
\end{table*}%

\section{The Details of LLMs Considered}
\label{sec:details}

Due to the involvement of different LLMs in constructing performance records for various benchmarks, the LLMs associated with each benchmark vary. Given that each benchmark involves thousands of LLMs and that table space is limited, we are unable to directly display the complete list of all LLMs. Please refer to our code for detailed information. In this section, we provide statistical results of these LLMs for reference.

On the one hand, we specifically show the number of LLMs involved in each benchmark in Table~\ref{tab:detail_routereval}. It should be noted that some performance records contain duplicates or errors, and the performance of some open-source models collected is very low (below 0.1), which has almost no reference value. We have carefully filtered these data. It can be seen that the number of LLMs involved in different benchmarks ranges from 1800 to 5000, with a total of more than 8,500 LLMs involved.

\begin{figure}[t]
  \includegraphics[width=\linewidth]{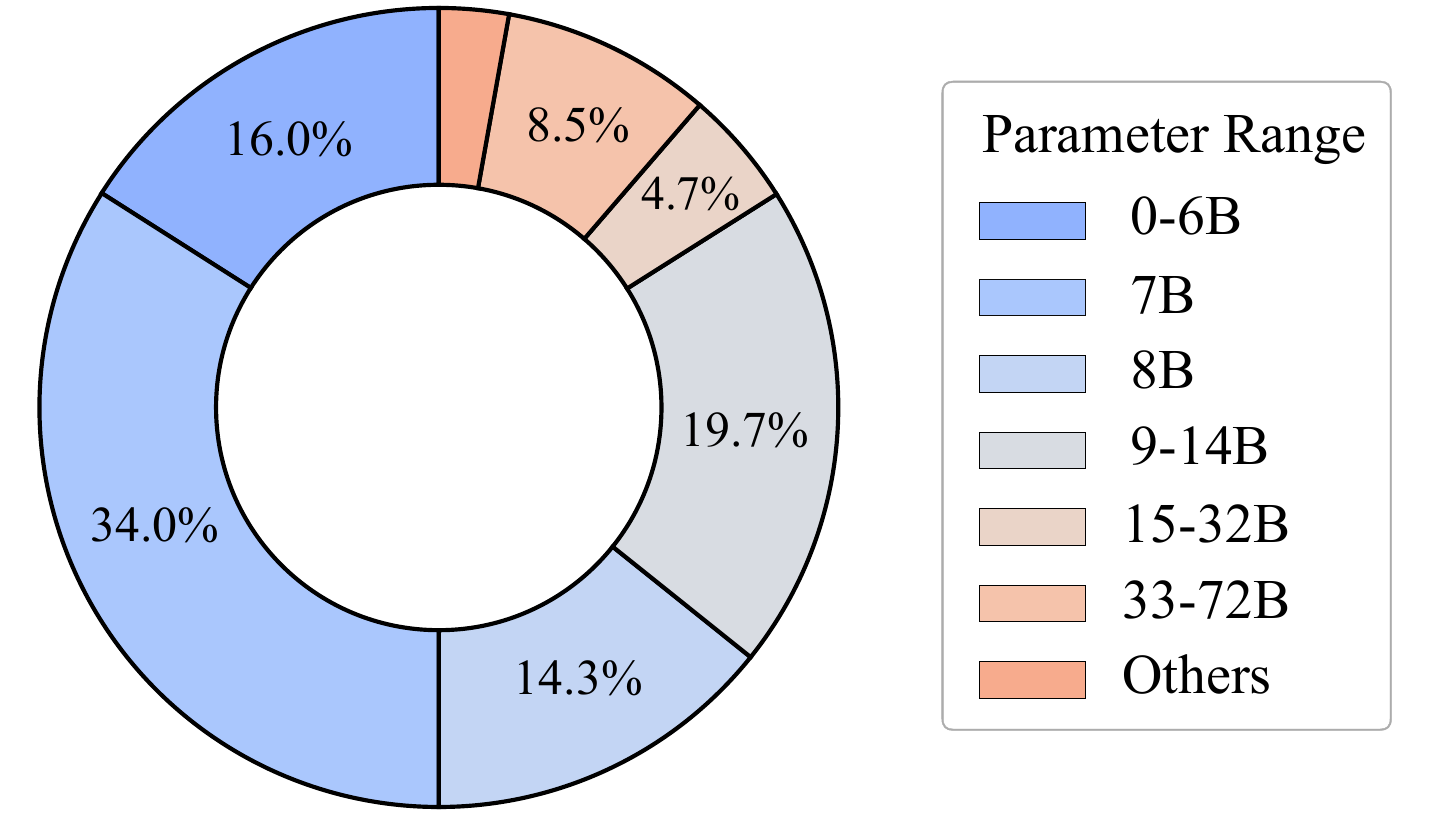}
  \caption{\textbf{The Distribution of Model Parameters}. We conduct a statistical analysis of the parameter counts for all LLMs considered in this paper and find that models with 7B parameters are predominant. }
  \label{fig:stat}
\end{figure}

\begin{figure*}[t]
  \includegraphics[width=\linewidth]{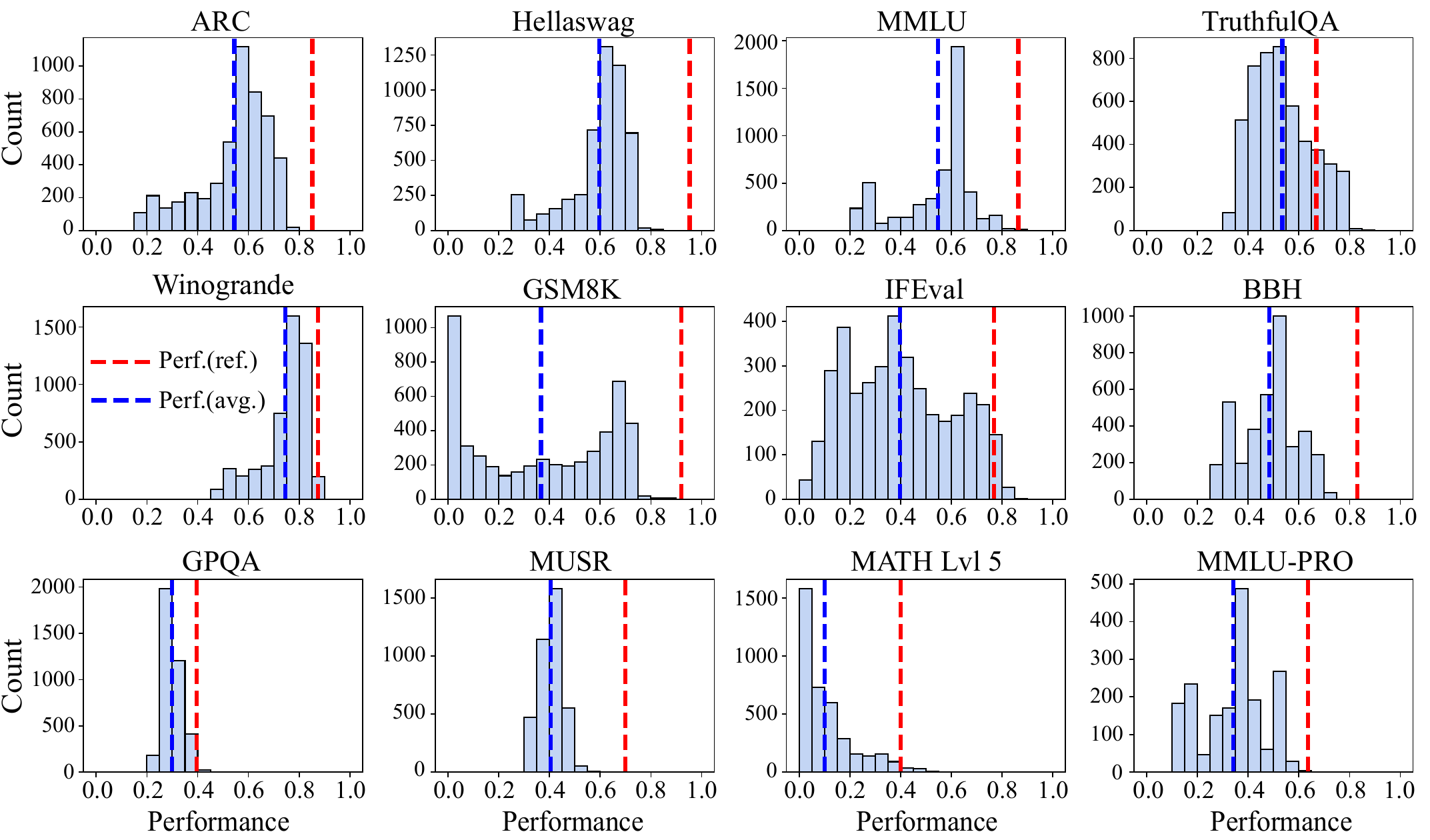}
  \caption{\textbf{The Distribution of Model Performance Under 12 Benchmarks}. We present the performance distribution of the LLMs involved in each of the 12 benchmarks. As shown in Fig.~\ref{fig:stat}, the majority of LLMs are 7B models, which tend to exhibit relatively weaker performance across the benchmarks from a statistical standpoint. }
  \label{fig:modelper}
\end{figure*}

Furthermore, we have conducted a statistical analysis of the parameters of all LLMs involved in this study, as shown in Fig.~\ref{fig:stat}. It can be observed that the majority of these LLMs are 7B models, most of which are relatively weak open-source models. Specifically, in Fig.~\ref{fig:modelper}, we present the performance distribution of these models across 12 different benchmarks. Nevertheless, the experiments in the main text have demonstrated that even using relatively weak models as candidates, the Routing LLMs paradigm can still fully leverage the strengths of each model, achieve complementary advantages, and ultimately achieve outstanding overall performance.

\section{Related Work}
\label{sec:related}

\subsection{Routing LLMs}

LLMs trained on massive datasets with trillions of tokens \citep{radford2019language,brown2020language} have transformed natural language processing and other research areas. However, deploying these models for specific tasks—such as classification or question-answering—presents distinct challenges. Traditional model selection techniques in statistics and machine learning, such as k-fold cross-validation, estimate population errors for in-distribution data \citep{bishop2006pattern,hastie2009elements,raschka2018model,Hendy:2023,Narayanan:2023}. Yet, these methods are impractical for LLMs due to their vast, often inaccessible training data and the computational infeasibility of retraining them repeatedly.

To address this, routing~\cite{frick2025prompt,zhang2025capability,zhang2023model,ramirez2023cache,mohammadshahi2024routoo,nguyen2024metallm,liu2024optllm,ramirez2024optimising,zhang2023ecoassistant} offers an efficient solution by dynamically selecting the most suitable LLM for a given input, eliminating the need to evaluate all possible models. Routing strategies fall into two main categories: non-predictive and predictive. Non-predictive approaches, like FrugalGPT \citep{chen2023frugalgpt}, sequentially invoke LLMs until an output meets a predefined quality threshold determined by a judger. Other non-predictive methods use layered frameworks to escalate complex queries to more advanced models \citep{tabi} or combine smaller models with LLMs \citep{automix,llmcascades,orchestrallm,routingtoexpert,HuBieLi2024}. 

In contrast, predictive routing leverages machine learning techniques—such as supervised learning \citep{llmbenchmark}, reward models \citep{tryage,routingtoexpert}, or meta-models predicting performance scores \citep{flyswat}—to preemptively identify the optimal LLM, thereby reducing latency and computational overhead.
Recent advancements have enhanced routing’s effectiveness through innovative techniques, including neural network-based meta-models \citep{Ding:2024,Sakota:2024,CheJiaLin2024,Aggarwal:2024}, k-nearest neighbors \citep{HuBieLi2024,Shnitzer:2023,Stripelis:2024,Lee:2024}, matrix factorization \citep{OngAlmWu2024,ZhuWuWen2024,Li:2025}, and graph neural networks \citep{Feng:2024}. These methods enable coordination among multiple LLMs or even sub-models within a single framework, such as MatFormer \citep{Devvrit:2024,Cai:2024}. 

Unlike earlier approaches that required scoring outputs from every candidate LLM \citep{liu2021simcls,ravaut2022summareranker,jiang2023llm}, modern routing uses benchmark data to map each LLM’s strengths across tasks and domains, requiring inference only from the chosen model. Further developments include supervised router training \citep{LuYuaLin2024,Zhao:2024} and efforts to improve robustness \citep{Dann:2024,Montreuil:2025,Shafran:2025}, highlighting routing’s ability to balance cost and performance.

Despite these innovations, the growing variety of routing strategies lacks a unified evaluation standard. The proposed RouterEval aims to bridge that gap by proposing a systematic framework to assess router effectiveness, providing a foundation for future advancements.

\subsection{Model Ensemble}

LLM ensemble methods offer a compelling strategy for leveraging the collective strengths of multiple LLMs to boost performance, efficiency, and robustness in NLP tasks. Departing from the conventional reliance on a single model for inference, LLM ensembles integrate outputs from diverse models, either by aggregating predictions after inference or by orchestrating their contributions throughout the process. Drawing inspiration from classical machine learning ensemble techniques—such as bagging and boosting—these methods are tailored to tackle the distinct challenges of modern LLMs, including their immense scale, computational demands, and architectural complexity.

A notable approach within LLM ensembling is the development of LLM cascades, designed to optimize inference efficiency without compromising output quality. For example, \citet{chen2023frugalgpt} propose a sequential framework in which LLMs are arranged in order of increasing parameter size. The process starts with a smaller, computationally lightweight model producing an initial output. If this output satisfies a predefined quality threshold, the cascade terminates, delivering the result; if not, the task escalates to progressively larger and more capable models. Similarly, \citet{yue2024large} introduce a verification-based cascade, where a smaller LLM generates a preliminary answer that is then evaluated for accuracy. Should it fall short, a more powerful LLM steps in to refine or correct the response. These cascading strategies significantly alleviate the computational load of depending exclusively on large-scale LLMs, making them especially valuable for resource-limited scenarios.

Another promising avenue in LLM ensemble research centers on generating a range of candidate outputs from distinct LLMs and then selecting or synthesizing the optimal result. This method capitalizes on the diversity of model architectures, training datasets, and reasoning abilities to elevate overall performance. For instance, \citet{lee2023ensemble} employ this technique to enhance instruction-tuning data construction, choosing the most effective instruction from a pool of candidates produced by various LLMs. In a similar vein, \citet{jiang2023llm} investigate unsupervised evaluation metrics—such as BERTScore \cite{Zhang2020BERTScore}, BLEURT \cite{sellam2020bleurt}, BARTScore \cite{yuan2021bartscore}, and scores derived from ChatGPT—to rank and select the strongest output from a set of candidates. However, their work reveals a key challenge: the effectiveness of this selection hinges on the quality and variety within the candidate pool. To address this limitation, \citet{jiang2023llm} propose an advanced fusion model that takes the highest-ranked candidates as inputs and generates a polished final output, seamlessly blending the strengths of individual predictions.

Overall, the LLMs ensemble paradigm, for a given input, emphasizes requiring all candidate LLMs to perform inference, followed by aggregating and organizing their outputs. The final result is typically determined through strategies like majority voting. In contrast, the Routing LLMs paradigm prioritizes efficiency by assigning tasks to specific candidates before inference, thereby reducing computational overhead and enhancing efficiency. Naturally, as these paradigms are not entirely distinct and can draw inspiration from one another, there remains a degree of technical overlap, particularly in various optimization techniques and improvement strategies.

\subsection{Scaling Law}

Scaling laws~\cite{li2025mis} have become a cornerstone for deciphering the behavior of deep learning models across a wide array of domains and tasks. They provide critical insights into how performance correlates with key factors such as dataset size, model capacity, and computational resources. Early work by \citet{banko2001scaling} laid the groundwork by identifying a power-law relationship between validation error and training dataset size in tasks like confusion set disambiguation. Their findings revealed that as the dataset grows, the average error decreases predictably, while the model size needed to effectively fit the data scales log-linearly. This seminal observation was later expanded by \citet{amodei2016deep}, who demonstrated power-law improvements in word error rate with increased training data for the 38M-parameter Deep Speech 2 model, and by \citet{hestness2017deep}, who extended these exponential trends to diverse fields such as machine translation, language modeling, image processing, and speech recognition. These studies collectively highlighted the robustness of scaling laws, showing that performance gains remain consistent even as models and architectures evolve.

Building on this foundation, subsequent research has pushed the boundaries of scale while refining the implications of these laws. For instance, \citet{kaplan2020scaling} investigated models with up to 1.5B parameters trained on 23B tokens, deriving power-law relationships to optimize computational budget allocation. However, later critiques from \citet{hoffmann2022training} and \citet{hu2024minicpm} pointed out an underestimation of required training data, underscoring subtle methodological challenges in scaling studies. Beyond sheer scale, researchers have delved into more nuanced phenomena: \citet{wei2022emergent} identified emergent abilities in large language models that are absent in smaller ones, while \citet{hernandez2021scaling} explored scaling laws in transfer learning and finetuning contexts. Architectural diversity has also come under scrutiny, with \citet{tay2022scaling} showing that not all model designs scale equally, advocating for scaling studies to inform architecture development.
The adaptability of scaling laws to emerging paradigms is evident in recent innovations. Hybrid models like Mamba \citep{gu2023mamba}, analyzed by \citet{poli2024mechanistic}, alongside specialized scaling laws for mixture-of-experts models \citep{clark2022unified,fedus2022switch,shazeer2017outrageously} and sparse architectures \citep{frantar2023scaling,zhu2017prune}, demonstrate their versatility. The scope of scaling laws extends far beyond language tasks, encompassing vision-language models \citep{cherti2023reproducible,henighan2020scaling}, reinforcement learning \citep{hilton2023scaling,gao2023scaling}, and recommendation systems \citep{ardalani2022understanding}, underscoring their wide-ranging applicability.

While Transformer-based models \citep{vaswani2017attention}—exemplified by behemoths like Llama 3 with 405B parameters—dominate scaling law research due to their exceptional scalability, alternative architectures have not been overlooked. For comparison, ResNet101 boasts a modest 44M parameters \citep{sorscher2022beyond}, where \citet{sorscher2022beyond} investigated data pruning laws. Smaller-scale studies have also employed MLPs or SVMs \citep{hashimoto2021model} to probe scaling behavior. Additional dimensions, such as the impact of data quality and language-specific effects on scaling coefficients \citep{bansal2022data,zhang2022examining}, as well as multimodal scaling in foundation models \citep{aghajanyan2023scaling}, further enrich this research landscape.

Together, these efforts illustrate that scaling laws do more than predict performance—they serve as a guiding framework for resource allocation, architecture design, and generalization across domains. As such, they have become an indispensable tool in advancing artificial intelligence research and its real-world deployment, offering a lens through which we can better understand and harness the potential of ever-growing models and datasets.

\subsection{Recommender System}

Recommender systems \cite{zhao2024recommender} have emerged as essential tools to tackle the challenge of information overload, offering tailored content and services to users across diverse online platforms \cite{wu2022disentangled, fan2022graph}. These systems primarily rely on two core methodologies: Collaborative Filtering and Content-based recommendation. Collaborative Filtering, the most prevalent approach, harnesses historical user-item interactions—such as purchase histories or ratings—to uncover behavioral similarities among users and forecast their future preferences \cite{fan2019deep_dscf}. A key technique within Collaborative Filtering, Matrix Factorization, transforms discrete user and item identities into continuous embedding vectors, facilitating efficient computation of recommendation scores \cite{fan2018deep, fan2019deep_daso, zhao2021autoloss, zhaok2021autoemb}. In contrast, content-based methods enhance precision by incorporating supplementary data, such as user demographics or item descriptions, with a special focus on widely available textual information~\cite{vasile2016meta}.

The rise of deep learning has revolutionized recommender systems, unlocking advanced representation learning capabilities \cite{fan2022comprehensive}. For example, Neural Matrix Factorization leverages deep neural networks  to model non-linear interactions between users and items, outperforming traditional inner product techniques \cite{he2017neural}. Meanwhile, Graph Neural Networks have gained traction by representing user-item relationships as graph-structured data, employing message propagation to generate insightful node embeddings \cite{ying2018graph, fan2020graph, ma2021deep, derr2020epidemic}. To incorporate textual insights, models like DeepCoNN utilize Convolutional Neural Networks to process user reviews, thereby improving rating predictions \cite{zheng2017joint}. Similarly, NARRE introduces a neural attention mechanism that not only predicts ratings but also delivers review-level explanations~\cite{chen2018neural}. These breakthroughs highlight the profound influence of deep learning on enhancing recommendation quality.

In recent developments, the incorporation of language models has propelled recommender systems to new heights by tapping into their ability to comprehend and generate human-like natural language \cite{wu2020mind, wu2023personalized, dongre2023deep}. For instance, BERT4Rec employs Bidirectional Encoder Representations from Transformers to capture the sequential patterns in user behavior, significantly improving sequential recommendation tasks \cite{sun2019bert4rec}. Likewise, transformer-based approaches, such as the framework by \cite{liu2023chatgpt}, leverage the generative power of Transformers to recommend items while simultaneously crafting explanatory narratives. These innovations enable highly contextualized and personalized recommendations, with applications ranging from news curation \cite{wu2020mind} to drug suggestions~\cite{dongre2023deep}. This convergence of natural language processing and recommender system design exemplifies their dynamic evolution, promising ever more sophisticated and user-centric solutions.

In essence, Routing LLMs can be regarded as a specialized subset of RS. In this framework, the input corresponds to the user in a traditional RS, the pool of LLM candidates represents the items to be recommended, and the performance record serves as the interaction history between users and items. The router’s task is to select an appropriate LLM from this pool based on the given input, aiming to optimize for various goals such as maximizing accuracy, minimizing computational cost, or reducing hallucination. Unlike conventional RS, however, Routing LLMs face a significant constraint: the "user information" available for collection is extremely limited, and annotating or labeling this data poses a considerable challenge. This scarcity of detailed input data complicates the routing process, requiring innovative approaches to achieve effective recommendations.

\subsection{LLM Model Fusion}

Model merging~\cite{lu2024merge} has become a pivotal technique in the realm of LLMs, enabling the integration of strengths from multiple pre-trained or fine-tuned models to boost performance, adaptability, and efficiency. This approach can be broadly divided into two paradigms: zero-shot merging, which fuses models without further training, and merge-then-train, which involves refining the combined model after integration. 

Early zero-shot techniques, such as weight averaging~\cite{nagarajan2021uniform,wortsman2022model} and Linear Mode Connectivity, laid the groundwork, evolving into more advanced methods like Task Arithmetic\cite{ilharco2023editing}—where task vectors steer parameter adjustments—and TIES\cite{yadav2023tiesmerging}, which reduces interference through trimming and conflict resolution. Recent innovations, including DARE\cite{yu2024language} and Evolutionary Model Merge\cite{akiba2024evolutionary}, further refine this by optimizing selective parameters or inference pathways, all without additional training. On the other hand, merge-then-train strategies, such as Fisher Merging\cite{matena2022merging}, utilize the Fisher information matrix to assign parameter weights, while RegMean\cite{jin2023dataless} fine-tunes linear merging on a per-layer basis, carefully balancing embeddings and biases. 
However, both approaches encounter difficulties when merging models with divergent initializations, prompting research into permutation symmetries~\cite{ainsworth2022git,verma2024merging} to better align parameters.
A notable distinction exists between model merging and \textit{model fusion}. Merging typically aims for efficiency by creating a single, cohesive model \cite{Singh:2020}, whereas fusion often combines multiple models to enhance quality, potentially at the expense of speed \cite{Ravaut:2022,Jiang:2023}. 

Within the merging domain, weighted-average techniques—refined by methods like Hessian-based estimates \cite{daheim2023model} or pruning-enhanced Fisher weights \cite{nathan2024fisher}—adjust parameter significance but may fail to capture task-specific subtleties, resulting in performance degradation (e.g., a reported 10\% drop with basic averaging \cite{task_arithmetic}). To counter this, the notion of \textit{task vectors}, defined as 
$\bm{\tau}_{t} = \bm{\theta}_{t}^{\text{ft}} - \bm{\theta}^{\text{pre}}$
 \cite{task_arithmetic}, has gained prominence. These vectors encapsulate task-specific shifts in parameter space, facilitating precise conflict resolution during merging. Building on this, methods like Task Arithmetic  \cite{task_arithmetic}, AdaMerging \cite{AdaMerging}, and TIES-Merging~\cite{ties} address redundancy and sign discrepancies, improving cross-model compatibility.

Resolving parameter conflicts remains a core challenge, inspiring a variety of innovative strategies.  Task Arithmetic \cite{task_arithmetic} introduced arithmetic-based vector merging, while  TIES-Merging \cite{ties} and  AdaMerging enhance this by targeting interference sources. Evolutionary methods \cite{evaluate_merge} blend TIES with optimized inference routes, and practical applications like  MetaGPT \cite{metagpt} and LLM evaluators \cite{Kim2024Prometheus2A} showcase real-world efficacy. Alternatively,  ZipIt \cite{zipit} retains correlated parameters while preserving distinct layers, offering adaptability. Additional advancements, such as geometric weight analysis \cite{10.1145/325334.325242,jang2024model} and safety alignment     \cite{hammoud2024model}, further enrich the field. For models with shared architectures and initializations, zero-shot merging stands out as a key focus, striking a balance between efficiency and performance without the computational burden of retraining. This makes it a foundational element in the ongoing evolution of LLMs.

\begin{table*}[t]
  \centering
  \resizebox*{0.99\linewidth}{!}{
    \begin{tabular}{lllll}
    \toprule
    \textbf{ $m$} & \textbf{Group} & \textbf{Oracle $r_o$ } & \textbf{$r_o(0.5)$} & \textbf{Details} \\
    \midrule
    3     & all-strong & 0.972 & 0.924 & [0.869, 0.878, 0.879] \\
    3     & all\_weak & 0.765 & 0.583 & [0.406, 0.392, 0.408] \\
    3     & strong-to-weak & 0.938 & 0.755 &  [0.241, 0.599, 0.878] \\
    5     & all-strong & 0.983 &  0.930 & [0.863, 0.869, 0.878, 0.879, 0.892] \\
    5     & all\_weak &  0.843 & 0.566 & [0.279, 0.300, 0.313, 0.282, 0.276] \\
    5     & strong-to-weak & 0.943 & 0.761 & [0.330, 0.462, 0.615, 0.622, 0.869] \\
    10    & all-strong & 0.984 & 0.881 & [0.766, 0.776, 0.781, 0.772, 0.771, 0.783, 0.787, 0.780, 0.778, 0.782] \\
    10    & all-weak & 0.955 & 0.620 & [0.256, 0.299, 0.276, 0.296, 0.284, 0.287, 0.274, 0.280, 0.301, 0.278] \\
    10    & strong-to-weak & 0.981 & 0.765 & [0.245, 0.259, 0.487, 0.534, 0.577, 0.614, 0.606, 0.628, 0.658, 0.869] \\
    100   & all-strong &  0.996 & 0.890 & See Fig.~\ref{fig:dist_can} \\
    100   & all-weak & 1.000 & 0.630 & See Fig.~\ref{fig:dist_can} \\
    100   & strong-to-weak & 1.000 & 0.769 & See Fig.~\ref{fig:dist_can} \\
    1000  & all-strong & 1.000 & 0.839 & See Fig.~\ref{fig:dist_can} \\
    1000  & all-weak & 1.000 &  0.642 & See Fig.~\ref{fig:dist_can} \\
    1000  & strong-to-weak & 1.000 & 0.769 & See Fig.~\ref{fig:dist_can} \\
    \bottomrule
    \end{tabular}%
    }
    \caption{\textbf{The Details of Candidate Groups on MMLU.} \( m \) denotes the number of LLMs per candidate group. "Details" shows the performance of each LLM in the corresponding group. When \( m = 100 \) or \( m = 1000 \), the number of candidates is too large to present individually. Therefore, we only display their performance distributions in Fig.~\ref{fig:dist_can}. For specific performance metrics and the details of other evaluations, please refer directly to our code. }
  \label{tab:det_group}%
\end{table*}%

\begin{figure*}[t]
  \includegraphics[width=\linewidth]{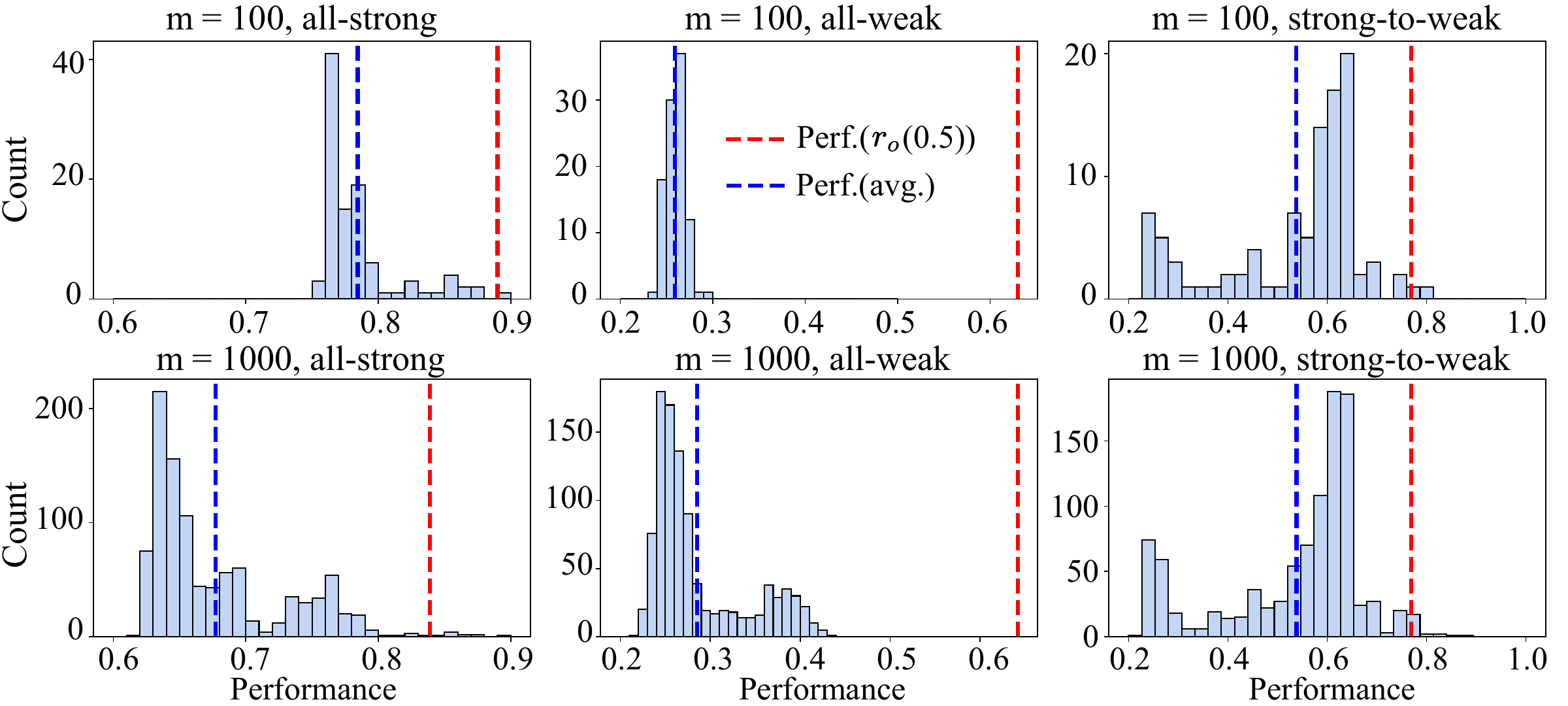}
  \caption{\textbf{The Distribution of Model Performance in Candidate Group on MMLU}. Due to the space limitations of Table \ref{tab:det_group}, it is not feasible to present the detailed performance of all candidates in the candidate groups where \( m = 100 \) or \( m = 1000 \). Therefore, we illustrate their performance distributions in this figure. For specific performance metrics and the details of other evaluations, please refer directly to our code.  }
  \label{fig:dist_can}
\end{figure*}

\subsection{Mixture-of-Experts (MoE)}
The MoE framework, first proposed as a technique for training independent models with distinct parameters and a routing mechanism \citep{Jacobs:1991, Jordan:1993}, has matured into a robust strategy for scaling neural networks. Originally, MoE models were designed to delegate tasks to specialized sub-models of uniform size, providing an appealing alternative to the limitations of single-model specialization \citep{Jang:2023, Douillard:2024}. With sub-models of equal capacity, the routing rule—responsible for directing inputs to the appropriate expert—did not need to account for varying computational costs. Over time, however, the paradigm has shifted toward integrating MoE into larger architectures, such as Transformers, where sub-models function as interconnected components within a cohesive system \citep{Fedus:2022, Zhou:2022}. This evolution has unlocked the power of sparse activation, where only a subset of parameters is engaged for each input, significantly boosting efficiency without sacrificing performance. A prime example is Mixtral \citep{jiang2024mixtral}, which competes with dense LLMs while activating far fewer parameters.

Modern MoE implementations have further refined this approach, with innovations like those from \citet{shazeer2017outrageously}, who introduced router networks to dynamically activate specific experts for individual input tokens. This technique has become a cornerstone of LLMs, celebrated for its generative capabilities and computational efficiency. Building on this, model mixture techniques have expanded the MoE framework by incorporating diverse dense LLM models—regardless of their size—into a unified system. For instance, Branch-Train-MiX \citep{sukhbaatar2024branchtrainmix} starts with a seed dense LLM, branches into parallel expert models during training, and later merges them into MoE layers by averaging non-expert parameters; yet, this method is limited to models sharing identical architectures. In contrast, model fusion strategies \citep{wan2024knowledge, wang2023fusing} blend expert outputs to harness insights from distinct data distributions. Most recently, UltraFuser \citep{ding2024mastering} has pushed the boundaries further with a token-level soft gating mechanism and a two-stage training process, offering enhanced flexibility in combining expert contributions. These developments highlight MoE’s remarkable ability to balance scalability, specialization, and resource efficiency, cementing its role as a pivotal advancement in contemporary machine learning research.  The Routing LLMs paradigm is a special type of MoE. Experts can be seen as LLMs in a candidate pool, and are selected by a router to process a given input. While MoE generally can choose multiple "experts", the current Routing LLMs paradigm only selects one LLM.

\section{The Details of Candidate Groups}
\label{sec:details_group}

In Section \ref{sec:constr}, we set three candidate types for the given benchmark and $m$, namely "all-strong," "all-weak," and "strong-to-weak." The final model performance is the average of the results from these three candidates. Specifically, we sort all $N$ LLMs with performance between 0.1 and 0.9 based on their individual performance on the given benchmark, obtaining $\{\ell_i^\prime\}_{i=1}^N$. We then consider the following optimization problem regarding the performance of $G$ and its corresponding oracle $r_o$, as shown in Eq.~(\ref{eq:opti2}),

\begin{equation}
\hat{G} = \max\nolimits_G \text{Perf.}(r_o, G).
\label{eq:opti2}
\end{equation}

When the $m$ LLMs in $G$ are all selected from $\{\ell_i^\prime\}_{i=1}^{\lfloor0.2N\rfloor}$ and $\{\ell_i^\prime\}_{i=\lfloor0.8N\rfloor}^{N}$ respectively, $\hat{G}$ forms the "all-strong" group and the "all-weak" group. Meanwhile, $\hat{G}$ forms the "strong-to-weak" group when the $j$-th LLM in $G$ is selected from $\{\ell_i^\prime\}_{i=(j-1)m}^{\min(jm,N)}$.

By solving Eq.~(\ref{eq:opti2}), we obtained the candidate selections for the three group types under the given benchmark and $m$. For example, in the case of MMLU, we detail these selections in Table \ref{tab:det_group}. Additionally, for the settings of $m=100$ and $m=1000$, due to space limitations, we only show their performance distributions in Fig.~\ref{fig:dist_can}. For specific candidates of these settings and other evaluation candidates, please refer directly to our code.

\section{The Reproducing Details of Baselines}
\label{sec:routers}

In this section, we provide the specific implementation methods for all the existing routers mentioned in Section \ref{sec:baseline}. Since some settings may not have been specifically discussed in their original papers, we attempt to supplement them. Specifically,

\begin{itemize}
    \item \textbf{LinearR}: A simple linear layer is used as the classifier. The input to the linear layer is the query representation, and the output dimension corresponds to the number of LLMs in the candidate pool, representing the selection scores for each LLM. As mentioned in Section \ref{sec:format}, we employ RoBERTa as the example encoder. The output dimension of RoBERTa is $768$, hence the input dimension of the linear layer is also $768$. During training, BCEWithLogitsLoss is used as the loss function with a batch size of $1$, learning rate of $1e-2$, and trained for $10$ epochs. During testing, the query representation is input to obtain an $m$-dimensional score vector, and the LLM with the highest score is selected for response generation.
    \item  \textbf{MLPR}: A Multi-Layer Perceptron (MLP) serves as the classifier, with the hidden layer size set to $256$. The input is the $768$-dimensional query representation, and the output is an $m$-dimensional score vector, consistent with LinearR. Training employs BCEWithLogitsLoss with a batch size of $1$, learning rate of $1e-4$, and runs for $100$ epochs.
    \item \textbf{MLC}: To handle the multi-label property of text embeddings, a multi-class classification(MLC) approach is used. Specifically, during training, the model treats multiple labels as positive classes for supervised learning. During  inference, the input embedding passes through the classifier to  generate a probability distribution, and select the highest one as the final router result. Training employs BCEWithLogitsLoss with a batch size of $10$, and runs for $10$ epochs.
    \item \textbf{C-RoBERTa}: We used K-Means to cluster the text embeddings of the training set into  $k$ clusters.  Specifically, for each cluster, the performance of each  model in the  candidate model pool is evaluated using the samples within that cluster. Then, the best-performing model is selected as the  dedicated predictor  for that cluster. During inference, the Euclidean  distance between the  test sample and each cluster center is calculated. The sample is matched to the nearest cluster using a nearest neighbor  strategy and the corresponding dedicated predictor is used as the router  result.  In this paper, $k$ is set to 3.
    \item \textbf{PRknn}: A k-Nearest Neighbors (KNN) classifier is adopted. During testing, given a test query representation, we retrieve the $K$ train queries with the smallest Euclidean distances. We then compute the average scores of the $m$ models on these $K$ train queries and select the model with the highest average score to process the test query. In our experiments, $K=5$.
\end{itemize}

\begin{table*}[t]
  \centering
  
  \resizebox*{0.99\linewidth}{!}{
    \begin{tabular}{|c|l|rrrr|rrrr|rrrr|rrrr|}
    \toprule
    \multicolumn{1}{|r}{} &       & \multicolumn{4}{c|}{\textbf{GPQA}}      & \multicolumn{4}{c|}{\textbf{MUSR}} & \multicolumn{4}{c|}{\textbf{MATH Lvl 5}}     & \multicolumn{4}{c|}{\textbf{MMLU-PRO}} \\
\cmidrule{3-18}    \multicolumn{1}{|r}{} & Router & \multicolumn{1}{l}{$\mu_o$↑} & \multicolumn{1}{l}{$V_R$↑} & \multicolumn{1}{l}{$V_B$↑} & \multicolumn{1}{l|}{$E_p$} & \multicolumn{1}{l}{$\mu_o$↑} & \multicolumn{1}{l}{$V_R$↑} & \multicolumn{1}{l}{$V_B$↑} & \multicolumn{1}{l|}{$E_p$} & \multicolumn{1}{l}{$\mu_o$↑} & \multicolumn{1}{l}{$V_R$↑} & \multicolumn{1}{l}{$V_B$↑} & \multicolumn{1}{l|}{$E_p$} & \multicolumn{1}{l}{$\mu_o$↑} & \multicolumn{1}{l}{$V_R$↑} & \multicolumn{1}{l}{$V_B$↑} & \multicolumn{1}{l|}{$E_p$} \\
    \midrule
    \multirow{8}[2]{*}{\begin{sideways}$m=10$\end{sideways}} & \cellcolor[rgb]{ .996,  .89,  .871}Oracle $r_o$ & \cellcolor[rgb]{ .996,  .89,  .871}0.93  & \cellcolor[rgb]{ .996,  .89,  .871}2.34  & \cellcolor[rgb]{ .996,  .89,  .871}2.58  & \cellcolor[rgb]{ .996,  .89,  .871}1.58  & \cellcolor[rgb]{ .996,  .89,  .871}0.88  & \cellcolor[rgb]{ .996,  .89,  .871}1.26  & \cellcolor[rgb]{ .996,  .89,  .871}1.99  & \cellcolor[rgb]{ .996,  .89,  .871}1.97  & \cellcolor[rgb]{ .996,  .89,  .871}0.71  & \cellcolor[rgb]{ .996,  .89,  .871}1.77  & \cellcolor[rgb]{ .996,  .89,  .871}1.96  & \cellcolor[rgb]{ .996,  .89,  .871}2.17  & \cellcolor[rgb]{ .996,  .89,  .871}0.81  & \cellcolor[rgb]{ .996,  .89,  .871}1.27  & \cellcolor[rgb]{ .996,  .89,  .871}2.73  & \cellcolor[rgb]{ .996,  .89,  .871}1.96  \\
          & \cellcolor[rgb]{ .996,  .89,  .871}$r_o$(0.5) & \cellcolor[rgb]{ .996,  .89,  .871}0.61  & \cellcolor[rgb]{ .996,  .89,  .871}1.54  & \cellcolor[rgb]{ .996,  .89,  .871}1.70  & \cellcolor[rgb]{ .996,  .89,  .871}2.92  & \cellcolor[rgb]{ .996,  .89,  .871}0.63  & \cellcolor[rgb]{ .996,  .89,  .871}0.90  & \cellcolor[rgb]{ .996,  .89,  .871}1.41  & \cellcolor[rgb]{ .996,  .89,  .871}3.01  & \cellcolor[rgb]{ .996,  .89,  .871}0.51  & \cellcolor[rgb]{ .996,  .89,  .871}1.28  & \cellcolor[rgb]{ .996,  .89,  .871}1.36  & \cellcolor[rgb]{ .996,  .89,  .871}3.04  & \cellcolor[rgb]{ .996,  .89,  .871}0.59  & \cellcolor[rgb]{ .996,  .89,  .871}0.92  & \cellcolor[rgb]{ .996,  .89,  .871}1.75  & \cellcolor[rgb]{ .996,  .89,  .871}2.98  \\
          & \cellcolor[rgb]{ .854, .909, .988}LinearR & \cellcolor[rgb]{ .854, .909, .988}\underline{\textbf{0.43}} & \cellcolor[rgb]{ .854, .909, .988}\underline{\textbf{1.09}} & \cellcolor[rgb]{ .854, .909, .988}\underline{\textbf{1.19}} & \cellcolor[rgb]{ .854, .909, .988}3.15  & \cellcolor[rgb]{ .854, .909, .988}\underline{\textbf{0.48}} & \cellcolor[rgb]{ .854, .909, .988}\underline{\textbf{0.68}} & \cellcolor[rgb]{ .854, .909, .988}\underline{\textbf{1.07}} & \cellcolor[rgb]{ .854, .909, .988}3.24  & \cellcolor[rgb]{ .854, .909, .988}0.39  & \cellcolor[rgb]{ .854, .909, .988}0.99  & \cellcolor[rgb]{ .854, .909, .988}1.00  & \cellcolor[rgb]{ .854, .909, .988}3.20  & \cellcolor[rgb]{ .854, .909, .988}\underline{\textbf{0.52}} & \cellcolor[rgb]{ .854, .909, .988}\underline{\textbf{0.82}} & \cellcolor[rgb]{ .854, .909, .988}\underline{\textbf{1.00}} & \cellcolor[rgb]{ .854, .909, .988}3.08  \\
          & \cellcolor[rgb]{ .854, .909, .988}MLPR & \cellcolor[rgb]{ .854, .909, .988}0.40  & \cellcolor[rgb]{ .854, .909, .988}1.01  & \cellcolor[rgb]{ .854, .909, .988}1.09  & \cellcolor[rgb]{ .854, .909, .988}3.10  & \cellcolor[rgb]{ .854, .909, .988}0.44  & \cellcolor[rgb]{ .854, .909, .988}0.63  & \cellcolor[rgb]{ .854, .909, .988}0.99  & \cellcolor[rgb]{ .854, .909, .988}3.27  & \cellcolor[rgb]{ .854, .909, .988}0.40  & \cellcolor[rgb]{ .854, .909, .988}0.99  & \cellcolor[rgb]{ .854, .909, .988}0.97  & \cellcolor[rgb]{ .854, .909, .988}3.17  & \cellcolor[rgb]{ .854, .909, .988}0.51  & \cellcolor[rgb]{ .854, .909, .988}0.81  & \cellcolor[rgb]{ .854, .909, .988}0.97  & \cellcolor[rgb]{ .854, .909, .988}3.07  \\
          & \cellcolor[rgb]{ .854, .909, .988}C-RoBERTa & \cellcolor[rgb]{ .854, .909, .988}0.34  & \cellcolor[rgb]{ .854, .909, .988}0.85  & \cellcolor[rgb]{ .854, .909, .988}0.90  & \cellcolor[rgb]{ .854, .909, .988}1.03  & \cellcolor[rgb]{ .854, .909, .988}0.41  & \cellcolor[rgb]{ .854, .909, .988}0.59  & \cellcolor[rgb]{ .854, .909, .988}0.91  & \cellcolor[rgb]{ .854, .909, .988}1.54  & \cellcolor[rgb]{ .854, .909, .988}\underline{\textbf{0.40}} & \cellcolor[rgb]{ .854, .909, .988}\underline{\textbf{0.99}} & \cellcolor[rgb]{ .854, .909, .988}\underline{\textbf{0.98}} & \cellcolor[rgb]{ .854, .909, .988}0.99  & \cellcolor[rgb]{ .854, .909, .988}0.52  & \cellcolor[rgb]{ .854, .909, .988}0.81  & \cellcolor[rgb]{ .854, .909, .988}1.00  & \cellcolor[rgb]{ .854, .909, .988}0.50  \\
          & \cellcolor[rgb]{ .854, .909, .988}MLC & \cellcolor[rgb]{ .854, .909, .988}0.36  & \cellcolor[rgb]{ .854, .909, .988}0.90  & \cellcolor[rgb]{ .854, .909, .988}0.97  & \cellcolor[rgb]{ .854, .909, .988}0.49  & \cellcolor[rgb]{ .854, .909, .988}0.39  & \cellcolor[rgb]{ .854, .909, .988}0.55  & \cellcolor[rgb]{ .854, .909, .988}0.86  & \cellcolor[rgb]{ .854, .909, .988}1.13  & \cellcolor[rgb]{ .854, .909, .988}0.31  & \cellcolor[rgb]{ .854, .909, .988}0.77  & \cellcolor[rgb]{ .854, .909, .988}0.76  & \cellcolor[rgb]{ .854, .909, .988}0.92  & \cellcolor[rgb]{ .854, .909, .988}0.52  & \cellcolor[rgb]{ .854, .909, .988}0.81  & \cellcolor[rgb]{ .854, .909, .988}0.99  & \cellcolor[rgb]{ .854, .909, .988}1.10  \\
          & \cellcolor[rgb]{ .854, .909, .988}PRknn & \cellcolor[rgb]{ .854, .909, .988}0.37  & \cellcolor[rgb]{ .854, .909, .988}0.93  & \cellcolor[rgb]{ .854, .909, .988}1.01  & \cellcolor[rgb]{ .854, .909, .988}3.29  & \cellcolor[rgb]{ .854, .909, .988}0.39  & \cellcolor[rgb]{ .854, .909, .988}0.56  & \cellcolor[rgb]{ .854, .909, .988}0.88  & \cellcolor[rgb]{ .854, .909, .988}3.30  & \cellcolor[rgb]{ .854, .909, .988}0.37  & \cellcolor[rgb]{ .854, .909, .988}0.93  & \cellcolor[rgb]{ .854, .909, .988}0.95  & \cellcolor[rgb]{ .854, .909, .988}3.30  & \cellcolor[rgb]{ .854, .909, .988}0.45  & \cellcolor[rgb]{ .854, .909, .988}0.70  & \cellcolor[rgb]{ .854, .909, .988}0.90  & \cellcolor[rgb]{ .854, .909, .988}3.30  \\
          & \cellcolor[rgb]{ .854, .909, .988}Random & \cellcolor[rgb]{ .854, .909, .988}0.30  & \cellcolor[rgb]{ .854, .909, .988}0.75  & \cellcolor[rgb]{ .854, .909, .988}0.82  & \cellcolor[rgb]{ .854, .909, .988}3.32  & \cellcolor[rgb]{ .854, .909, .988}0.38  & \cellcolor[rgb]{ .854, .909, .988}0.54  & \cellcolor[rgb]{ .854, .909, .988}0.84  & \cellcolor[rgb]{ .854, .909, .988}3.32  & \cellcolor[rgb]{ .854, .909, .988}0.32  & \cellcolor[rgb]{ .854, .909, .988}0.79  & \cellcolor[rgb]{ .854, .909, .988}0.77  & \cellcolor[rgb]{ .854, .909, .988}3.32  & \cellcolor[rgb]{ .854, .909, .988}0.36  & \cellcolor[rgb]{ .854, .909, .988}0.57  & \cellcolor[rgb]{ .854, .909, .988}0.76  & \cellcolor[rgb]{ .854, .909, .988}3.32  \\
    \midrule
    \multirow{8}[2]{*}{\begin{sideways}$m=100$\end{sideways}} & \cellcolor[rgb]{ .996,  .89,  .871}Oracle $r_o$ & \cellcolor[rgb]{ .996,  .89,  .871}0.99  & \cellcolor[rgb]{ .996,  .89,  .871}2.50  & \cellcolor[rgb]{ .996,  .89,  .871}2.54  & \cellcolor[rgb]{ .996,  .89,  .871}4.62  & \cellcolor[rgb]{ .996,  .89,  .871}0.96  & \cellcolor[rgb]{ .996,  .89,  .871}1.37  & \cellcolor[rgb]{ .996,  .89,  .871}2.06  & \cellcolor[rgb]{ .996,  .89,  .871}4.67  & \cellcolor[rgb]{ .996,  .89,  .871}0.89  & \cellcolor[rgb]{ .996,  .89,  .871}2.22  & \cellcolor[rgb]{ .996,  .89,  .871}2.40  & \cellcolor[rgb]{ .996,  .89,  .871}4.13  & \cellcolor[rgb]{ .996,  .89,  .871}0.97  & \cellcolor[rgb]{ .996,  .89,  .871}1.53  & \cellcolor[rgb]{ .996,  .89,  .871}3.60  & \cellcolor[rgb]{ .996,  .89,  .871}4.23  \\
          & \cellcolor[rgb]{ .996,  .89,  .871}$r_o$(0.5) & \cellcolor[rgb]{ .996,  .89,  .871}0.64  & \cellcolor[rgb]{ .996,  .89,  .871}1.62  & \cellcolor[rgb]{ .996,  .89,  .871}1.63  & \cellcolor[rgb]{ .996,  .89,  .871}6.15  & \cellcolor[rgb]{ .996,  .89,  .871}0.66  & \cellcolor[rgb]{ .996,  .89,  .871}0.94  & \cellcolor[rgb]{ .996,  .89,  .871}1.41  & \cellcolor[rgb]{ .996,  .89,  .871}6.12  & \cellcolor[rgb]{ .996,  .89,  .871}0.58  & \cellcolor[rgb]{ .996,  .89,  .871}1.45  & \cellcolor[rgb]{ .996,  .89,  .871}1.53  & \cellcolor[rgb]{ .996,  .89,  .871}5.90  & \cellcolor[rgb]{ .996,  .89,  .871}0.64  & \cellcolor[rgb]{ .996,  .89,  .871}1.01  & \cellcolor[rgb]{ .996,  .89,  .871}2.18  & \cellcolor[rgb]{ .996,  .89,  .871}5.97  \\
          & \cellcolor[rgb]{ .854, .909, .988}LinearR & \cellcolor[rgb]{ .854, .909, .988}\underline{\textbf{0.43}} & \cellcolor[rgb]{ .854, .909, .988}\underline{\textbf{1.09}} & \cellcolor[rgb]{ .854, .909, .988}\underline{\textbf{1.08}} & \cellcolor[rgb]{ .854, .909, .988}6.49  & \cellcolor[rgb]{ .854, .909, .988}0.48  & \cellcolor[rgb]{ .854, .909, .988}0.68  & \cellcolor[rgb]{ .854, .909, .988}0.99  & \cellcolor[rgb]{ .854, .909, .988}6.55  & \cellcolor[rgb]{ .854, .909, .988}0.40  & \cellcolor[rgb]{ .854, .909, .988}1.01  & \cellcolor[rgb]{ .854, .909, .988}0.94  & \cellcolor[rgb]{ .854, .909, .988}6.52  & \cellcolor[rgb]{ .854, .909, .988}0.46  & \cellcolor[rgb]{ .854, .909, .988}0.72  & \cellcolor[rgb]{ .854, .909, .988}0.95  & \cellcolor[rgb]{ .854, .909, .988}6.50  \\
          & \cellcolor[rgb]{ .854, .909, .988}MLPR & \cellcolor[rgb]{ .854, .909, .988}0.38  & \cellcolor[rgb]{ .854, .909, .988}0.94  & \cellcolor[rgb]{ .854, .909, .988}0.93  & \cellcolor[rgb]{ .854, .909, .988}6.58  & \cellcolor[rgb]{ .854, .909, .988}0.45  & \cellcolor[rgb]{ .854, .909, .988}0.65  & \cellcolor[rgb]{ .854, .909, .988}0.94  & \cellcolor[rgb]{ .854, .909, .988}6.59  & \cellcolor[rgb]{ .854, .909, .988}\underline{\textbf{0.41}} & \cellcolor[rgb]{ .854, .909, .988}\underline{\textbf{1.04}} & \cellcolor[rgb]{ .854, .909, .988}\underline{\textbf{1.00}} & \cellcolor[rgb]{ .854, .909, .988}6.49  & \cellcolor[rgb]{ .854, .909, .988}\underline{\textbf{0.46}} & \cellcolor[rgb]{ .854, .909, .988}\underline{\textbf{0.72}} & \cellcolor[rgb]{ .854, .909, .988}\underline{\textbf{0.97}} & \cellcolor[rgb]{ .854, .909, .988}6.53  \\
          & \cellcolor[rgb]{ .854, .909, .988}C-RoBERTa & \cellcolor[rgb]{ .854, .909, .988}0.35  & \cellcolor[rgb]{ .854, .909, .988}0.88  & \cellcolor[rgb]{ .854, .909, .988}0.88  & \cellcolor[rgb]{ .854, .909, .988}1.55  & \cellcolor[rgb]{ .854, .909, .988}\underline{\textbf{0.49}} & \cellcolor[rgb]{ .854, .909, .988}\underline{\textbf{0.70}} & \cellcolor[rgb]{ .854, .909, .988}\underline{\textbf{1.02}} & \cellcolor[rgb]{ .854, .909, .988}1.54  & \cellcolor[rgb]{ .854, .909, .988}0.41  & \cellcolor[rgb]{ .854, .909, .988}1.03  & \cellcolor[rgb]{ .854, .909, .988}0.98  & \cellcolor[rgb]{ .854, .909, .988}1.15  & \cellcolor[rgb]{ .854, .909, .988}0.45  & \cellcolor[rgb]{ .854, .909, .988}0.71  & \cellcolor[rgb]{ .854, .909, .988}0.93  & \cellcolor[rgb]{ .854, .909, .988}1.00  \\
          & \cellcolor[rgb]{ .854, .909, .988}MLC & \cellcolor[rgb]{ .854, .909, .988}0.29  & \cellcolor[rgb]{ .854, .909, .988}0.72  & \cellcolor[rgb]{ .854, .909, .988}0.70  & \cellcolor[rgb]{ .854, .909, .988}1.06  & \cellcolor[rgb]{ .854, .909, .988}0.40  & \cellcolor[rgb]{ .854, .909, .988}0.57  & \cellcolor[rgb]{ .854, .909, .988}0.83  & \cellcolor[rgb]{ .854, .909, .988}3.33  & \cellcolor[rgb]{ .854, .909, .988}0.38  & \cellcolor[rgb]{ .854, .909, .988}0.94  & \cellcolor[rgb]{ .854, .909, .988}0.88  & \cellcolor[rgb]{ .854, .909, .988}0.00  & \cellcolor[rgb]{ .854, .909, .988}0.45  & \cellcolor[rgb]{ .854, .909, .988}0.70  & \cellcolor[rgb]{ .854, .909, .988}0.94  & \cellcolor[rgb]{ .854, .909, .988}2.81  \\
          & \cellcolor[rgb]{ .854, .909, .988}PRknn & \cellcolor[rgb]{ .854, .909, .988}0.43  & \cellcolor[rgb]{ .854, .909, .988}1.07  & \cellcolor[rgb]{ .854, .909, .988}1.04  & \cellcolor[rgb]{ .854, .909, .988}6.61  & \cellcolor[rgb]{ .854, .909, .988}0.38  & \cellcolor[rgb]{ .854, .909, .988}0.54  & \cellcolor[rgb]{ .854, .909, .988}0.78  & \cellcolor[rgb]{ .854, .909, .988}6.62  & \cellcolor[rgb]{ .854, .909, .988}0.36  & \cellcolor[rgb]{ .854, .909, .988}0.89  & \cellcolor[rgb]{ .854, .909, .988}0.88  & \cellcolor[rgb]{ .854, .909, .988}6.62  & \cellcolor[rgb]{ .854, .909, .988}0.36  & \cellcolor[rgb]{ .854, .909, .988}0.56  & \cellcolor[rgb]{ .854, .909, .988}0.83  & \cellcolor[rgb]{ .854, .909, .988}6.62  \\
          & \cellcolor[rgb]{ .854, .909, .988}Random & \cellcolor[rgb]{ .854, .909, .988}0.29  & \cellcolor[rgb]{ .854, .909, .988}0.74  & \cellcolor[rgb]{ .854, .909, .988}0.73  & \cellcolor[rgb]{ .854, .909, .988}6.64  & \cellcolor[rgb]{ .854, .909, .988}0.36  & \cellcolor[rgb]{ .854, .909, .988}0.51  & \cellcolor[rgb]{ .854, .909, .988}0.75  & \cellcolor[rgb]{ .854, .909, .988}6.64  & \cellcolor[rgb]{ .854, .909, .988}0.27  & \cellcolor[rgb]{ .854, .909, .988}0.67  & \cellcolor[rgb]{ .854, .909, .988}0.67  & \cellcolor[rgb]{ .854, .909, .988}6.64  & \cellcolor[rgb]{ .854, .909, .988}0.32  & \cellcolor[rgb]{ .854, .909, .988}0.50  & \cellcolor[rgb]{ .854, .909, .988}0.75  & \cellcolor[rgb]{ .854, .909, .988}6.64  \\
    \midrule
    \multirow{8}[2]{*}{\begin{sideways}$m=1000$\end{sideways}} & \cellcolor[rgb]{ .996,  .89,  .871}Oracle $r_o$ & \cellcolor[rgb]{ .996,  .89,  .871}1.00  & \cellcolor[rgb]{ .996,  .89,  .871}2.52  & \cellcolor[rgb]{ .996,  .89,  .871}2.34  & \cellcolor[rgb]{ .996,  .89,  .871}7.95  & \cellcolor[rgb]{ .996,  .89,  .871}0.99  & \cellcolor[rgb]{ .996,  .89,  .871}1.41  & \cellcolor[rgb]{ .996,  .89,  .871}1.92  & \cellcolor[rgb]{ .996,  .89,  .871}7.66  & \cellcolor[rgb]{ .996,  .89,  .871}0.97  & \cellcolor[rgb]{ .996,  .89,  .871}2.41  & \cellcolor[rgb]{ .996,  .89,  .871}2.16  & \cellcolor[rgb]{ .996,  .89,  .871}6.53  & \cellcolor[rgb]{ .996,  .89,  .871}0.99  & \cellcolor[rgb]{ .996,  .89,  .871}1.55  & \cellcolor[rgb]{ .996,  .89,  .871}1.77  & \cellcolor[rgb]{ .996,  .89,  .871}7.66  \\
          & \cellcolor[rgb]{ .996,  .89,  .871}$r_o$(0.5) & \cellcolor[rgb]{ .996,  .89,  .871}0.65  & \cellcolor[rgb]{ .996,  .89,  .871}1.64  & \cellcolor[rgb]{ .996,  .89,  .871}1.53  & \cellcolor[rgb]{ .996,  .89,  .871}9.47  & \cellcolor[rgb]{ .996,  .89,  .871}0.67  & \cellcolor[rgb]{ .996,  .89,  .871}0.96  & \cellcolor[rgb]{ .996,  .89,  .871}1.31  & \cellcolor[rgb]{ .996,  .89,  .871}9.30  & \cellcolor[rgb]{ .996,  .89,  .871}0.60  & \cellcolor[rgb]{ .996,  .89,  .871}1.49  & \cellcolor[rgb]{ .996,  .89,  .871}1.32  & \cellcolor[rgb]{ .996,  .89,  .871}8.87  & \cellcolor[rgb]{ .996,  .89,  .871}0.67  & \cellcolor[rgb]{ .996,  .89,  .871}1.05  & \cellcolor[rgb]{ .996,  .89,  .871}1.18  & \cellcolor[rgb]{ .996,  .89,  .871}9.32  \\
          & \cellcolor[rgb]{ .854, .909, .988}LinearR & \cellcolor[rgb]{ .854, .909, .988}\underline{\textbf{0.47}} & \cellcolor[rgb]{ .854, .909, .988}\underline{\textbf{1.18}} & \cellcolor[rgb]{ .854, .909, .988}\underline{\textbf{1.08}} & \cellcolor[rgb]{ .854, .909, .988}9.81  & \cellcolor[rgb]{ .854, .909, .988}\underline{\textbf{0.54}} & \cellcolor[rgb]{ .854, .909, .988}\underline{\textbf{0.77}} & \cellcolor[rgb]{ .854, .909, .988}\underline{\textbf{1.03}} & \cellcolor[rgb]{ .854, .909, .988}9.87  & \cellcolor[rgb]{ .854, .909, .988}0.46  & \cellcolor[rgb]{ .854, .909, .988}1.15  & \cellcolor[rgb]{ .854, .909, .988}0.91  & \cellcolor[rgb]{ .854, .909, .988}9.77  & \cellcolor[rgb]{ .854, .909, .988}\underline{\textbf{0.62}} & \cellcolor[rgb]{ .854, .909, .988}\underline{\textbf{0.98}} & \cellcolor[rgb]{ .854, .909, .988}\underline{\textbf{1.04}} & \cellcolor[rgb]{ .854, .909, .988}9.76  \\
          & \cellcolor[rgb]{ .854, .909, .988}MLPR & \cellcolor[rgb]{ .854, .909, .988}0.38  & \cellcolor[rgb]{ .854, .909, .988}0.94  & \cellcolor[rgb]{ .854, .909, .988}0.87  & \cellcolor[rgb]{ .854, .909, .988}9.88  & \cellcolor[rgb]{ .854, .909, .988}0.54  & \cellcolor[rgb]{ .854, .909, .988}0.77  & \cellcolor[rgb]{ .854, .909, .988}1.01  & \cellcolor[rgb]{ .854, .909, .988}9.92  & \cellcolor[rgb]{ .854, .909, .988}0.46  & \cellcolor[rgb]{ .854, .909, .988}1.15  & \cellcolor[rgb]{ .854, .909, .988}0.90  & \cellcolor[rgb]{ .854, .909, .988}9.72  & \cellcolor[rgb]{ .854, .909, .988}0.62  & \cellcolor[rgb]{ .854, .909, .988}0.97  & \cellcolor[rgb]{ .854, .909, .988}1.02  & \cellcolor[rgb]{ .854, .909, .988}9.77  \\
          & \cellcolor[rgb]{ .854, .909, .988}C-RoBERTa & \cellcolor[rgb]{ .854, .909, .988}0.40  & \cellcolor[rgb]{ .854, .909, .988}1.00  & \cellcolor[rgb]{ .854, .909, .988}0.91  & \cellcolor[rgb]{ .854, .909, .988}0.91  & \cellcolor[rgb]{ .854, .909, .988}0.53  & \cellcolor[rgb]{ .854, .909, .988}0.75  & \cellcolor[rgb]{ .854, .909, .988}0.98  & \cellcolor[rgb]{ .854, .909, .988}1.54  & \cellcolor[rgb]{ .854, .909, .988}\underline{\textbf{0.46}} & \cellcolor[rgb]{ .854, .909, .988}\underline{\textbf{1.16}} & \cellcolor[rgb]{ .854, .909, .988}\underline{\textbf{0.92}} & \cellcolor[rgb]{ .854, .909, .988}1.43  & \cellcolor[rgb]{ .854, .909, .988}0.61  & \cellcolor[rgb]{ .854, .909, .988}0.96  & \cellcolor[rgb]{ .854, .909, .988}1.00  & \cellcolor[rgb]{ .854, .909, .988}0.50  \\
          & \cellcolor[rgb]{ .854, .909, .988}MLC & \cellcolor[rgb]{ .854, .909, .988}0.28  & \cellcolor[rgb]{ .854, .909, .988}0.70  & \cellcolor[rgb]{ .854, .909, .988}0.64  & \cellcolor[rgb]{ .854, .909, .988}0.86  & \cellcolor[rgb]{ .854, .909, .988}0.49  & \cellcolor[rgb]{ .854, .909, .988}0.70  & \cellcolor[rgb]{ .854, .909, .988}0.92  & \cellcolor[rgb]{ .854, .909, .988}5.13  & \cellcolor[rgb]{ .854, .909, .988}0.41  & \cellcolor[rgb]{ .854, .909, .988}1.04  & \cellcolor[rgb]{ .854, .909, .988}0.77  & \cellcolor[rgb]{ .854, .909, .988}1.77  & \cellcolor[rgb]{ .854, .909, .988}0.52  & \cellcolor[rgb]{ .854, .909, .988}0.81  & \cellcolor[rgb]{ .854, .909, .988}0.76  & \cellcolor[rgb]{ .854, .909, .988}5.24  \\
          & \cellcolor[rgb]{ .854, .909, .988}PRknn & \cellcolor[rgb]{ .854, .909, .988}0.40  & \cellcolor[rgb]{ .854, .909, .988}1.01  & \cellcolor[rgb]{ .854, .909, .988}0.93  & \cellcolor[rgb]{ .854, .909, .988}9.94  & \cellcolor[rgb]{ .854, .909, .988}0.29  & \cellcolor[rgb]{ .854, .909, .988}0.41  & \cellcolor[rgb]{ .854, .909, .988}0.53  & \cellcolor[rgb]{ .854, .909, .988}9.94  & \cellcolor[rgb]{ .854, .909, .988}0.28  & \cellcolor[rgb]{ .854, .909, .988}0.70  & \cellcolor[rgb]{ .854, .909, .988}0.59  & \cellcolor[rgb]{ .854, .909, .988}9.94  & \cellcolor[rgb]{ .854, .909, .988}0.40  & \cellcolor[rgb]{ .854, .909, .988}0.63  & \cellcolor[rgb]{ .854, .909, .988}0.67  & \cellcolor[rgb]{ .854, .909, .988}9.94  \\
          & \cellcolor[rgb]{ .854, .909, .988}Random & \cellcolor[rgb]{ .854, .909, .988}0.31  & \cellcolor[rgb]{ .854, .909, .988}0.77  & \cellcolor[rgb]{ .854, .909, .988}0.71  & \cellcolor[rgb]{ .854, .909, .988}9.97  & \cellcolor[rgb]{ .854, .909, .988}0.36  & \cellcolor[rgb]{ .854, .909, .988}0.51  & \cellcolor[rgb]{ .854, .909, .988}0.69  & \cellcolor[rgb]{ .854, .909, .988}9.97  & \cellcolor[rgb]{ .854, .909, .988}0.23  & \cellcolor[rgb]{ .854, .909, .988}0.57  & \cellcolor[rgb]{ .854, .909, .988}0.48  & \cellcolor[rgb]{ .854, .909, .988}9.97  & \cellcolor[rgb]{ .854, .909, .988}0.35  & \cellcolor[rgb]{ .854, .909, .988}0.55  & \cellcolor[rgb]{ .854, .909, .988}0.58  & \cellcolor[rgb]{ .854, .909, .988}9.97  \\
    \bottomrule
    \end{tabular}%
    }
         \caption{\textbf{The Results on Hard Level RouterEval (part1)}. See Section \ref{sec:metrics} for detials of various metrics.  Red area and blue area highlights indicate the "Strong router" and "Existing router" mentioned in Section \ref{sec:baseline}, respectively. The best results in existing router methods are highlighted with underlines and on bold. The values in the table are rounded to two decimal places.}
  \label{tab:hard2}%
\end{table*}%

\section{The Results on Hard Level Settings}
\label{sec:res_hard}

In the main text, we primarily present the experimental results of RouterEval at the easy level, that is, the cases where \( m \in \{3, 5\} \). In this section, we provide the experimental results of RouterEval at the hard level, where \( m \in \{10, 100, 1000\} \). Given the scarcity of data, the difficulty of the classification problem at the hard level is significantly higher than that at the easy level. However, as demonstrated by the model-level scaling up phenomenon shown in the main text, the Routing LLMs paradigm can only exhibit its surprisingly strong performance when there are a sufficient number of LLM candidates, typically ranging from 100 to 1000 candidates. Therefore, exploring RouterEval at the hard level is highly necessary. Tables \ref{tab:hard1} and \ref{tab:hard2} show the classification performance of different routers, and the results are consistent with our analysis. Thus, the Routing LLMs paradigm still has considerable room for improvement.

\begin{table*}[t]
  \centering

  \resizebox*{0.99\linewidth}{!}{
\begin{tabular}{|c|l|rrrr|rrrr|rrrr|rrrr|}
    \toprule
    \multicolumn{1}{|r}{} &       & \multicolumn{4}{c|}{\textbf{ARC}} & \multicolumn{4}{c|}{\textbf{HellaSwag}} & \multicolumn{4}{c|}{\textbf{MMLU}} & \multicolumn{4}{c|}{\textbf{TruthfulQA}} \\
\cmidrule{3-18}    \multicolumn{1}{|r}{} & Router & \multicolumn{1}{l}{$\mu_o$↑} & \multicolumn{1}{l}{$V_R$↑} & \multicolumn{1}{l}{$V_B$↑} & \multicolumn{1}{l|}{$E_p$} & \multicolumn{1}{l}{$\mu_o$↑} & \multicolumn{1}{l}{$V_R$↑} & \multicolumn{1}{l}{$V_B$↑} & \multicolumn{1}{l|}{$E_p$} & \multicolumn{1}{l}{$\mu_o$↑} & \multicolumn{1}{l}{$V_R$↑} & \multicolumn{1}{l}{$V_B$↑} & \multicolumn{1}{l|}{$E_p$} & \multicolumn{1}{l}{$\mu_o$↑} & \multicolumn{1}{l}{$V_R$↑} & \multicolumn{1}{l}{$V_B$↑} & \multicolumn{1}{l|}{$E_p$} \\
    \midrule
    \multirow{8}[2]{*}{\begin{sideways}$m=10$\end{sideways}} & \cellcolor[rgb]{ .996,  .89,  .871}Oracle $r_o$ & \cellcolor[rgb]{ .996,  .89,  .871}0.90  & \cellcolor[rgb]{ .996,  .89,  .871}1.06  & \cellcolor[rgb]{ .996,  .89,  .871}1.48  & \cellcolor[rgb]{ .996,  .89,  .871}2.30  & \cellcolor[rgb]{ .996,  .89,  .871}0.84  & \cellcolor[rgb]{ .996,  .89,  .871}0.88  & \cellcolor[rgb]{ .996,  .89,  .871}1.15  & \cellcolor[rgb]{ .996,  .89,  .871}2.83  & \cellcolor[rgb]{ .996,  .89,  .871}0.97  & \cellcolor[rgb]{ .996,  .89,  .871}1.13  & \cellcolor[rgb]{ .996,  .89,  .871}1.85  & \cellcolor[rgb]{ .996,  .89,  .871}2.17  & \cellcolor[rgb]{ .996,  .89,  .871}0.92  & \cellcolor[rgb]{ .996,  .89,  .871}1.38  & \cellcolor[rgb]{ .996,  .89,  .871}1.28  & \cellcolor[rgb]{ .996,  .89,  .871}2.61  \\
\bottomrule
    \end{tabular}%
    }
     \caption{\textbf{The Results on Hard Level RouterEval (part2)}. See Section \ref{sec:metrics} for detials of various metrics.  Red area and blue area highlights indicate the "Strong router" and "Existing router" mentioned in Section \ref{sec:baseline}, respectively. The best results in existing router methods are highlighted with underlines and on bold. The values in the table are rounded to two decimal places.}
  \label{tab:hard1}%
\end{table*}%

\section{More Examples Visualization of Model-level Scaling Up}
\label{sec:more}


In Section \ref{sec:poten}, we discussed the model-level scaling up phenomenon on four well-known LLM benchmarks. In this section, we supplement our findings with observations from eight additional LLM benchmarks to illustrate the prevalence of this phenomenon. Specifically, similar to Section \ref{sec:poten}, we construct the oracle router \( r_o \) based on the performance record, and then define \( r_o(p) \) to create other routers with different capabilities as follows:
\begin{equation}
    r_o(p) = \begin{cases}
r_o, & \text{with probability } p, \\
\omega_m, & \text{with probability } 1-p,
\end{cases}
\end{equation}
where \( \omega_m \) is a router that samples uniformly from the \( m \) candidate LLMs with probability \( 1/m \). As \( p \to 1 \), \( r_o(p) \) approaches the oracle router \( r_o \), leading to the strongest classification performance among the \( m \) LLM candidates. Conversely, as \( p \to 0 \), \( r_o(p) \) degenerates into a random sampler.

\begin{figure*}[t]
 \centering
  \includegraphics[width=0.92\linewidth]{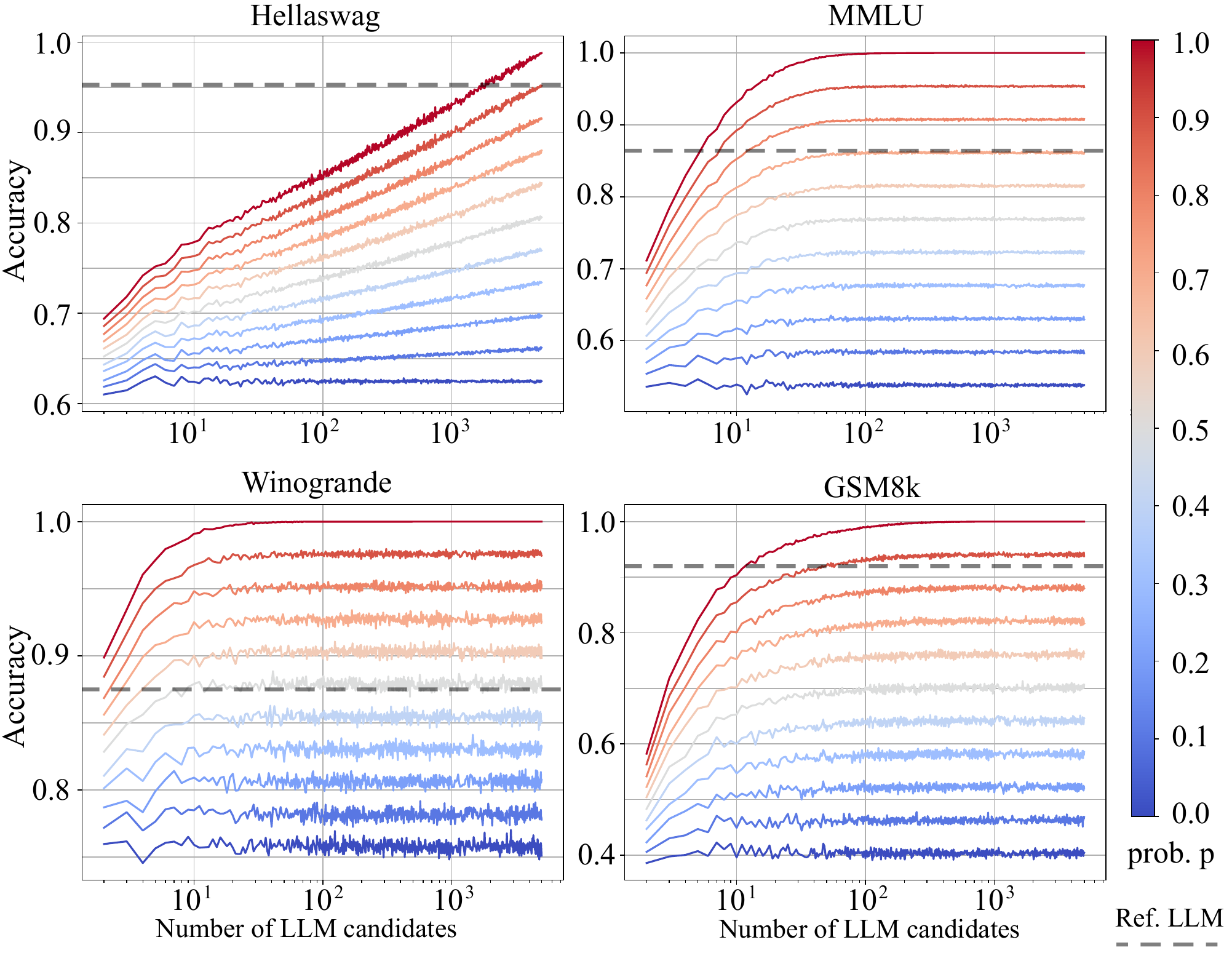}
  \includegraphics[width=0.92\linewidth]{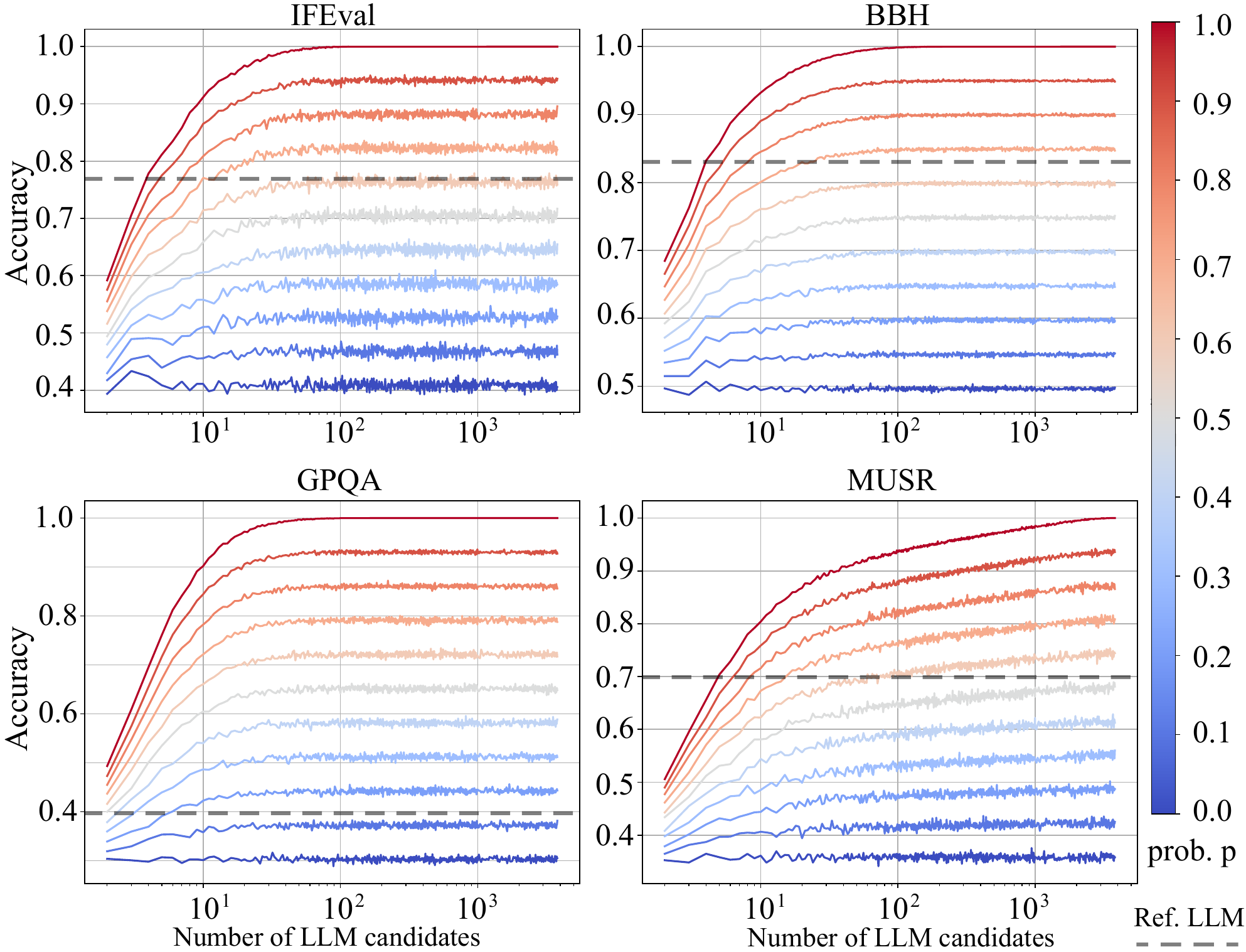}
  \caption{\textbf{The Model-level Scaling Up Phenomenon in Routing LLMs (part2)}. As shown in Section \ref{sec:poten}, the Prob. $p$ indicates the performance of the router, with values closer to 1 representing greater similarity to the oracle router's capability. If $p \to 0$, then $r_o(p)$ degenerates into a random sampler. When the router $r_o(p)$ reaches a certain level of capability, it induces a scaling up phenomenon in the Routing LLMs paradigm. Specifically, as the number of LLM candidates increases, performance rapidly improves. "Ref. LLM" denotes a representative LLM with strong performance on given benchmark, such as GPT-4.  }
  \label{fig:scaling2}
\end{figure*}

From the experimental results in Fig.~\ref{fig:scaling2}, we can draw similar observations. On the one hand, across different benchmarks, we still observe that with the support of a capable router, the overall capability of the model increases as the number of LLM candidates grows. On the other hand, even weak candidates can achieve satisfactory performance under the routing LLMs paradigm, working together to deliver good results, even when their number is relatively small, such as in the case of 3 to 10 LLMs.

\end{document}